%% file: main.tex
\documentclass[11pt]{article}
 
\usepackage[top=1in,bottom=1in,left=1in,right=1in]{geometry}
\usepackage[numbers]{natbib}

\input{formality.tex}

\input{macros.tex}

\title{
  Stochastic Graph Bandit Learning with Side-Observations
}

\usepackage{authblk}
\author[*]{Xueping Gong}
\author[$\dagger$]{Jiheng Zhang}

\affil[*$\dagger$]{Department of Industrial Engineering and Decision Analytics}
\affil[*$\dagger$]{The Hong Kong University of Science and Technology}

\date{}

\begin{document}
\maketitle

\begin{abstract}
  In this paper, we investigate the stochastic contextual bandit with general function space and graph feedback. 
  We propose an algorithm that addresses this problem by adapting to both the underlying graph structures and reward gaps. 
  To the best of our knowledge, our algorithm is the first to provide a gap-dependent upper bound in this stochastic setting, 
  bridging the research gap left by the work in \cite{contextualGB}. 
  In comparison to \cite{linGB_stochastic,linGB_adv,contextualGB}, our method offers improved regret upper bounds and does not require knowledge of graphical quantities.
  We conduct numerical experiments to demonstrate the effectiveness of our approach in terms of regret upper bounds.
  These findings highlight the significance of our algorithm in advancing the field of stochastic contextual bandits with graph feedback, opening up avenues for practical applications in various domains.
\end{abstract}

\input{intro.tex}

\input{setting.tex}

\input{method.tex}

\input{numerical.tex}

\input{conclusion.tex}

\newpage

\bibliographystyle{plain}
\bibliography{ref}

\newpage
\appendix

\input{appendix.tex}

\end{document}

%% file: formality.tex
\usepackage{algorithm}
\usepackage{algorithmicx}
\usepackage{amsmath,amssymb,amsfonts}
\usepackage{graphicx,color}
\usepackage{url}
\usepackage{subfiles}
\usepackage{booktabs}
\usepackage{mathtools}
\usepackage{microtype}
\usepackage[english]{babel}
\usepackage{caption}
\usepackage[toc,page,title,titletoc]{appendix}

\usepackage{tikz, pgfplots, pgfplotstable, subcaption}
\usepgfplotslibrary{fillbetween}
\usepackage{hyperref}       
\hypersetup{
  colorlinks=true,
  linkcolor=black,
  urlcolor=black,
  citecolor=black
}
\usepackage[noend]{algpseudocode}
\usepackage{setspace}
\usepackage{yhmath}
\usepackage[figurename=Fig.,font={small,stretch=0.84}]{caption}
\usepackage{multirow}

\usepackage{cleveref}
\crefname{section}{Section}{Sections}
\crefname{figure}{Figure}{Figures}
\crefname{mythm}{Theorem}{Theorems}
\crefname{mylem}{Lemma}{Lemmas}
\crefname{myrem}{Remark}{Remarks}
\crefname{appendix}{Appendix}{Appendicies}
\crefname{myprop}{Proposition}{Propositions}
\crefname{equation}{Equation}{Equations}

\usetikzlibrary{arrows.meta, calc, topaths, positioning, automata}
\pgfplotsset{compat=1.16}
\newenvironment{customlegend}[1][]{%
\begingroup
\csname pgfplots@init@cleared@structures\endcsname
\pgfplotsset{#1}%
}{%
\csname pgfplots@createlegend\endcsname
\endgroup
}%
\def\addlegendimage{\csname pgfplots@addlegendimage\endcsname}
\def\BibTeX{{\rm B\kern-.05em{\sc i\kern-.025em b}\kern-.08em
    T\kern-.1667em\lower.7ex\hbox{E}\kern-.125emX}}

%% file: macros.tex
\newcommand{\prob}[1]{\mathbb{P}( #1 )}

\newcommand{\mean}[1]{\mathbb{E} [ #1 ]}

\newcommand{\ind}[1]{ \mathbf{I} \left\{#1 \right\} }
\newcommand{\argmax}[0]{ \mathop{\arg\max}  }
\newcommand{\argmin}[0]{ \mathop{\arg\min}  }

\newtheorem{mythm}{\bf{Theorem}}[section]
\newtheorem{mylem}{\bf{Lemma}}[section]
\newtheorem{myprop}{\bf{Proposition}}[section]
\newtheorem{mydef}{\bf{Definition}}[section]
\newtheorem{mycol}{\bf{Corollary}}[section]
\newtheorem{myrem}{\bf{Remark}}[section]
\newtheorem{myasp}{\bf{Assumption}}[section]
\newenvironment{myproof}{ {\noindent\it Proof.\ }}{\hfill $\square$\par}

%% file: intro.tex
\section{Introduction}

The bandit framework has garnered significant attention from the online learning community due to its widespread applicability in diverse fields such as recommendation systems, portfolio selection, and clinical trials \cite{contextualBanditRecommendation}. 
Among the significant aspects of sequential decision making within this framework are side observations, 
which can be feedback from multiple sources \cite{valSideObservation} or contextual knowledge about the environment \cite{improvedUCB,CBwithPredictableRewards}. 
These are typically represented as graph feedback and contextual bandits respectively.



The multi-armed bandits framework with feedback graphs has emerged as a mature approach, 
providing a solid theoretical foundation for incorporating additional feedback into the exploration strategy \cite{MAB-GB,sideObservations,beyondBandits}. 
The contextual bandit problem is another well-established framework for decision-making under uncertainty \cite{banditBook,linContextual,improvedUCB}. 
Despite the considerable attention given to non-contextual bandits with feedback graphs, 
the exploration of contextual bandits with feedback graphs has been limited \cite{contextualGB,linGB_adv,linGB_stochastic}. 
\cite{linGB_adv,linGB_stochastic} mainly focus on linear payoffs in the adversarial and stochastic settings, respectively.
Notably, \cite{contextualGB} presented the first solution for general feedback graphs and function classes in the semi-adversarial settings, 
providing a minimax upper bound on regrets. 
However, to the best of our knowledge, there is no prior work considering stochastic feedback graphs and contexts while providing gap-dependent upper bounds for general reward functions and feedback graphs.

In this paper, we fill this gap by proposing a practical graph learning algorithm that can simultaneously adapt to time-varying graph structures and reward gaps. 
We introduce a probabilistic sampling method that focuses on informative actions discovered through graph structures and empirical evaluation.
Importantly, our method does not require prior knowledge of difficult-to-obtain graph parameters like independence numbers.
We also show the minimax optimality of our algorithm (up to logarithmic terms in the time horizon).
These theoretical results provide strong evidence that our method can effectively handle both easy and difficult bandit instances, ensuring consistently good performance across a wide range of scenarios.

The contributions of this paper are as follows:
\begin{enumerate}
\item We introduce a novel algorithm for contextual bandits with a general reward function space and directed feedback graphs.
We provide a gap-dependent upper bound on regrets, 
addressing a significant research gap in previous works \cite{contextualGB,linGB_adv}. 
Our algorithm exhibits adaptability to time-varying feedback graphs and contextual bandit instances, 
making it well-suited for stochastic settings.

\item Besides the improved regret bounds on easy instances, 
we further highlight the novelty of our algorithm through enhanced implementation practicability. 
Unlike existing methods, it does not require prior knowledge of difficult-to-obtain graph parameters, 
simplifying the implementation process \cite{linGB_adv}.
It also provides advantages in terms of offline regression oracles compared to methods relying on complex online regression oracles \cite{contextualGB}.

\item Extensive numerical experiments are conducted to validate the findings.
Our ablation experiments necessitate the special algorithmic design in graph feedback setting 
and showcase the adaptability of our approach by testing on different types of graphs, including a real-world dataset.
Further experiments demonstrate that the regret of the proposed method scales with the graph parameters rather than action set sizes.
\end{enumerate}

\subsection{Related Work}

Our work is closely related to contextual bandits, which have been extensively studied due to their wide range of applications, 
including scheduling, dynamic pricing, packet routing, online auctions, e-commerce, and matching markets \cite{MLsurvey}. 
Several formulations of contextual bandits have been proposed, 
such as linear bandits \cite{improvedUCB}, generalized linear bandits \cite{GLbandit,linContextual}, and kernelized bandits \cite{kernelisedContextualBandits,EfficientKernelUCB}. 
Researchers have designed algorithms tailored to specific function space structures, 
aiming to achieve near-optimal regret rates. 
A comprehensive summary of bandit algorithms can be found in the book by \cite{banditBook}. 
More recently, a line of works \cite{CBwithOracle,CBwithPredictableRewards,CBwithRegressionOracles,gong2023provably} has achieved optimal regret bounds for contextual bandits with a general function class, 
assuming access to a regression oracle. 
These findings are novel, requiring only minimal realizability assumptions. 
Our work builds upon these advancements and extends them to the setting of graph feedback.

Our work also relates to online learning with side information modeled by feedback graphs. 
The concept of feedback graphs was first introduced by \cite{valSideObservation} as a way to interpolate between full information settings, 
where the agent is aware of all rewards at each round, 
and bandit settings, where only the reward of the selected action is known. 
Online learning with feedback graphs has been extensively analyzed by \cite{beyondBandits} and other authors \cite{expG+GB,understandingGB,feedbackGraphsWithout,gangbandit,asymGB}. 
Various methods, including UCB \cite{GB_UCB_TS,sideObservations}, TS \cite{GB_UCB_TS}, EXP \cite{expG+GB,understandingGB,beyondBandits,feedbackGraphsWithout}, IDS \cite{IDSforGB}, 
and their variants, have been designed for this setting. 
Notably, \cite{RLfeedbackgraph} extended the graph feedback setting to the reinforcement learning. 
While the setting of graph feedback can be framed within the broader context of partial monitoring games \cite{banditBook}, 
algorithms for general partial monitoring games are not efficient for dealing with graph feedback due to the requirement of infinite or exponentially large feedback matrices.


Our proposed approach also offers significant advancements in the field of contextual bandits with graph feedback.
\cite{linGB_stochastic} investigate contextual bandits with side-observations and a linear payoff reward function.
Their objective is to design recommendation algorithms for users connected via social networks.
They propose a UCB-type learning algorithm, which provides gap-dependent upper bounds (Corollary 1 in \cite{linGB_stochastic}).
In our work, \cref{alg: ADA-G} and \cref{thm: gap-dependent upper bound} can be easily extended to infinite function spaces $\mathcal{F}$ by standard learning-theoretic complexity measures such as metric entropy. 
Furthermore, it is important to highlight that \cite{linGB_stochastic} deals exclusively with fixed feedback graphs, 
while our work considers time-varying graphs which contains fixed graphs as a special case.
On the other hand, the method proposed in \cite{linGB_stochastic} cannot be generalized to time-varying graphs, 
while our approach is designed to adapt to time-varying graphs.
Additionally, their method do not include the minimax optimality,
indicating that their method may perform worse on more difficult instances.

Similarly, the paper \cite{linGB_adv} focuses on adversarial linear contextual bandits with graph feedbacks.
They propose EXP-based algorithms that leverage both contexts and graph structures to guarantee the minimax regrets. 
\cite{linGB_adv} require knowledge of the independence number at each round to adapt to the time-varying graphs. 
They rely on these quantities to tune the input learning rate before the learning process starts. 
As \cite{contextualGB} point out, \cite{linGB_adv} ``assumes several unrealistic assumptions on both the policy class
and the context space''.
Actually, computing the independence number for a given graph is an NP problem \cite{NPindependence}, 
making this approach impractical.

The work of \cite{contextualGB} presents the first solution to contextual bandits with general function classes and graph feedback, 
resulting in the minimax regrets of $\tilde{\mathcal{O}}(\sqrt{\alpha T \log |\mathcal{F}|})$,
where $\alpha$ represents the uniform upper bound for each feedback graph. 
Their approach tackles this problem by reducing it to an online regression oracle.
However, designing and implementing online oracles are more challenging than offline oracles due to their sensitivity to the order of data sequence. 
In practice, the simple least square estimators in \cite{linGB_adv,linGB_stochastic} can serve as valid offline oracles but not valid online oracles. 
Finally, we emphasize that \cite{contextualGB}, as well as \cite{linGB_adv}, 
fail to generalize to gap-dependent upper bounds, 
which is the significant contribution of our work. 
Adapting the sampling probability to specific bandit instances and time-varying graphs would require non-trivial modifications to the algorithmic design and proofs.


%% file: setting.tex
\section{Problem Formulation}
\label{sec: setting}

Throughout this paper, we let $[n]$ denote the set $\{1, 2, \cdots , n\}$ for any positive integer $n$. 
We consider the following contextual bandits problem with informed feedback graphs. The learning process goes in $T$ rounds. 
At each round $t \in [T]$, the environment independently selects a context $x_t \in \mathcal{X} $ and 
a directed feedback graph $G_t \in \mathcal{G}$, 
where $\mathcal{X}$ is the domain of contexts and $\mathcal{G}$ is the collections of directed strongly observable graphs.
Both $G_t$ and $x_t$ are revealed to the learner at the beginning of each round $t$. 
Let $\mathcal{A}$ be the action set, which consists of nodes of graphs in $\mathcal{G}$.
For each $a\in \mathcal{A}$, denote the in-neighborhood of a node (arm) $a$ in $\mathcal{A}$ as $\mathcal{N}^{in}_a(G) =\{ v\in \mathcal{A}: (v,a) \in E \text{ for }G = (\mathcal{A},E)  \}$.
and the out-neighborhood of $a$ as $\mathcal{N}^{out}_a(G) =\{ v\in \mathcal{A}: (a,v) \in E \text{ for }G = (|\mathcal{A}|,E)  \}$.
Then the learner selects one of the actions $a_t \in \mathcal{A}$ and then observes rewards according to the feedback graph $G_t$. 
Specifically, for the action $a_t$, the learner observes the rewards of all actions in $\mathcal{N}^{out}_{a_t}(G_t)$, i.e., 
$\{(x_t,a,y_{t,a})\}$ for $a\in \mathcal{N}^{out}_{a_t}(G_t)$.
We assume all rewards are bounded in $[0,1]$ and all reward distributions are independent.

We further assume that the learner has access to a class of reward functions $\mathcal{F}\subset \mathcal{X}\times \mathcal{A} \to [0,1]$ (e.g., linear function classes) 
that characterizes the mean of the reward distribution for a given context-action pair.
In particular, we make the following standard realizability assumption \cite{fasterCB,CBwithRegressionOracles,instanceCB_RL}.
\begin{myasp}[realizability]
  \label{asp: realizability}
  There exists a function $f^*\in\mathcal{F}$ such that $f^*(x_t, a) = E[y_{t,a} | x_t]$ for all $a\in\mathcal{A}$ and all $t\in[T]$.
\end{myasp}
We define the induced policy of a function $f$ as $\pi_f(x) = \argmax_{a\in\mathcal{A}}f(x,a)  $.
The set of all induced policies, denoted as $\Pi = \{\pi_f | f \in \mathcal{F}\}$, forms the whole policy space.
The objective of the agent is to achieve low regret with respect to the optimal policy $\pi_{f^*}$,
and the regret over time horizon $T$ is defined as follows:
$$
Reg(T) = \sum_{t=1}^T   f^*(x_t, \pi_{f^*} (x_t)) - f^*(x_t, a_t)  .
$$
Here, $a_t$ is the selected action at the round $t$.

We further assume access to an offline least square regression oracle for function class $\mathcal{F}$. 
Based on the dataset $\{(x_n,a,y_{n,a})|a\in \mathcal{N}^{out}_{a_n}(G_n)\}_{n=1}^{t-1}$, 
the goal of the oracle is to find an estimator $\hat{f}_t\in\mathcal{F}$ via minimizing the cumulative square errors: 
$$
\hat{f}_t = \argmin_{f\in\mathcal{F}} \sum_{n=1}^{t-1} \sum_{a\in\mathcal{N}^{out}_{a_n}(G_n)}   ( f(x_n,a) -y_{n,a} )^2 .
$$
Compared to online oracles, which rely on the specific order of dataset inputs, 
designing algorithms with offline oracles is more efficient and straightforward \cite{efficientoracle}. 
Online oracles ensure empirical square loss through algorithmic design, 
while offline oracles automatically establish concentration results due to the independent and identically distributed (i.i.d.) nature of the data.
In this graph feedback setting, we summarize corresponding concentration properties in \cref{lem: high-probability events}.

To establish an upper bound that depends on the bandit gaps, 
we rely on a standard uniform gap assumption commonly used in stochastic settings.

\begin{myasp}[Uniform Gap Assumption]
\label{asp: uniform gap}
For all $x\in\mathcal{X}$, there exists a positive gap $\Delta$ such that
$$
f^*(x,\pi_{f^*}(x)) - f^*(x,a) \geq \Delta, \forall a\neq \pi_{f^*}(x).
$$
\end{myasp}

Next, we introduce some notations related to graph parameters, 
as they play a crucial role in controlling the regret order within this framework.

\begin{mydef}
An independence set of a graph $G$ is a subset of nodes in which no two distinct nodes are connected. The independence number of a graph refers to the size of its largest independence set. We denote the independence number as $\alpha(G)$.
\end{mydef}

Understanding and leveraging the independence number of a graph is essential for regulating the regret behavior in this setting.


%% file: method.tex
\section{Algorithm Design and Regret Analysis}
\label{sec: method}

\subsection{Algorithm design}
We present our method in \cref{alg: ADA-G},
where the side-observations, including contextual information and graph feedbacks,
are used nearly optimal.
In more detail, we operate in a doubling epoch schedule. 
Letting $\tau_m = 2^m$ with $\tau_0 = 0$, each epoch $m \geq 1$ consists of rounds $\tau_{m-1} + 1, . . . , \tau_m$, 
and there are $\lceil \log_2 T \rceil$ epochs in total.
To understand the motivation behind the our designed algorithm,
we point out several adaptivity features of our algorithms.

\begin{algorithm}[htbp]
	\renewcommand{\algorithmicrequire}{\textbf{Input:}}
	\renewcommand{\algorithmicensure}{\textbf{Output:}}
	\caption{An Adaptive Contextual Bandit algorithm with Graph feedback (AdaCB.G)}
	\label{alg: ADA-G}
	\begin{algorithmic}[1]
      \Require time horizon $T$, confidence parameter $\delta$, tuning parameters $\eta$ 
      \State Set epoch schedule $\{\tau_m=2^m,\forall m \in \mathbb{N} \}$ and the sample splitting schedule $t_m=\frac{\tau_m+\tau_{m-1}}{2}$
      \For{ epoch $m = 1,2,\cdots,\lceil \log_2 T \rceil$ }
      \State Compute the confidence radius $\beta_m = 16(\log T - m + 1) \log(2|\mathcal{F}||\mathcal{A}|^2T^2/\delta) $ 
      \State Compute the smoothing parameter $\mu_m = 64\log(4\delta^{-1}\log T) / (\tau_m -\tau_{m-1})$
      \State Compute the function 
      $$
      \hat{f}_m = \argmin_{f\in\mathcal{F}} \sum_{n=1}^{\tau_{m-1}} \sum_{a\in\mathcal{N}^{out}_{a_n}(G_n)}   ( f(x_n,a) -y_{n,a} )^2 
      $$ 
      via the \textbf{Offline Least Square Oracle}
      \State Compute $\mathcal{F}_m$ according to \eqref{eq: function space Fm}
      \State Compute the instance-dependent scale factor 
      $$
      \lambda_m = \frac{ \mathbb{E}_{x\sim\mathcal{D}_m}[ \ind{  \mathcal{A}(x;\mathcal{F}_m)>1  }   ] +\mu_m  }{ \sqrt{ \mathbb{E}_{x\sim\mathcal{D}_{m-1}}[ \ind{  \mathcal{A}(x;\mathcal{F}_{m-1})>1  }   ] +\mu_{m-1}   }    },
      $$
      where $\mathcal{D}_m\sim unif(x_{t_{m-1}+1},\cdots,x_{\tau_{m-1}})$ (for the first epoch, $\lambda_1=1$)
       
      \For{ round $t = \tau_{m-1}+1,\cdots, \tau_{m}$ }
            \State Observe the context $x_t$ and the graph $G_t$
            \State Compute the best arm candidate set $\mathcal{A}(x_t;\mathcal{F}_m)$
            \State Call the subroutine \textbf{ConstructExplorationSet} to find the exploration set $S_t$
            \State Compute $\gamma_t = \lambda_m \sqrt{  \frac{\eta |S_t| (\tau_{m-1}-\tau_{m-2})  }{  2 \log (   2\delta^{-1}|\mathcal{A}||\mathcal{F}| T^2   )  }     }    $ (for the first epoch, $\gamma_t=0$)
            \State Compute the sampling probabilities $p_t^*(a)$ according to \eqref{eq: sampling probability}
            \State Sample $a_t \sim p_t(\cdot)$ and take the action $a_t$
            \State Observe a feedback graph $\{ (a,y_{t,a})| a\in \mathcal{N}^{out}_{a_t}(G_t)\}$ from $G_t$
            \EndFor
         \EndFor
   \end{algorithmic}  
\end{algorithm}

\textbf{Adaptive to function spaces.} 
At each epoch $m$, \cref{alg: ADA-G} identifies the best empirical estimator $\hat{f}_m$ using historical data and offline least square oracle. 
The algorithm maintains a function space $\mathcal{F}_m$ defined by an upper confidence bound:
\begin{equation}
\label{eq: function space Fm}
\begin{array}{lcl}
\mathcal{F}_m = \Bigg\{ f\in\mathcal{F} |  \sum_{n=1}^{t_{m-1}} \sum_{a\in\mathcal{N}^{out}_{a_n}(G_n)}   ( f(x_n,a) -y_{n,a} )^2 \\ 
 \leq  \min_{\tilde{f}\in\mathcal{F}} \sum_{n=1}^{t_{m-1}} \sum_{a\in\mathcal{N}^{out}_{a_n}(G_n)}   ( \tilde{f}(x_n,a) -y_{n,a} )^2 +\beta_m  \Bigg\}
\end{array}
\end{equation}
which is the set of all plausible predictors that cannot yet be eliminated based on square loss confidence bounds.
It can be proved in \cref{lem: high-probability events} that the choice of $\beta_m$ guarantees
$
f^* \in \mathcal{F}_m, \text{ for } m=1,2,\cdots, \lceil \log_2 T \rceil,
$
with probability at least $1-\delta$.
To ensure the independence of $\mathcal{F}_m$ and $\lambda_m$, we split the dataset into two parts. 
The data from $1$ to $t_{m-1}$ is used to determine $\mathcal{F}_m$, 
while the data from $t_{m-1} + 1$ to $\tau_{m-1}$ is utilized to estimate the parameter $\lambda_m$. 
This separation enables the algorithm to maintain the independence between these two components, 
simplifying the regret analysis.

\textbf{Adaptive to action sets.} 
Using the obtained $\mathcal{F}_m$, we compute the data-driven candidate action set as
$$
    \mathcal{A}(x;\mathcal{F}_m) = \left\{a| a = \argmax_{a\in\mathcal{A}} f(x,a)  \text{ for some $f$ in $\mathcal{F}_m$}  \right\}.
$$
A systematic method of computing the candidate action set $\mathcal{A}(x;\mathcal{F}_m)$ is provided in Section 4 and Appendix A of \cite{instanceCB_RL}.
Our main objective is to ensure that each arm within $\mathcal{A}(x;\mathcal{F}_m)$ is observed sufficiently often. 
The additional graph feedback enables us to gather information on certain arms by playing adjacent arms in the graph. 
Leveraging this property of feedback graphs, 
we restrict exploration to a subset of arms while still obtaining sufficient information about all the arms in $\mathcal{A}(x;\mathcal{F}_m)$.

The exploration set $S_t$ is constructed by the subroutine \textbf{ConstructExplorationSet}, 
which is used by \cref{alg: ADA-G} to compute the sampling probabilities on each arm. 
  We modify the idea of constructing an independence set in \cite{contextualGB,expG+GB} to the induced subgraphs of $G_t$ for the arms in $\mathcal{A}(x_t;\mathcal{F}_m)$.
  It begins by sorting the arms in ascending order based on their gap estimates.
  Subsequently, $S_t$ is constructed greedily by iteratively selecting the arm with the smallest gap and removing its neighbors in $G_t$ from consideration.
  Since the chosen arms are non-adjacent in $G_t$, this procedure yields an independence set.
  The size of $S_t$ is thus bounded by the independence number $\alpha(G_t)$.
  Intuitively, by prioritizing arms with smaller estimated gaps, the subroutine efficiently explores the most promising regions of the action space while respecting the dependence structure in $G_t$, as imposed by its independence set construction.
  
\begin{figure}[h]
 \centering
\begin{tikzpicture}[shorten >=1pt,auto,node distance=4cm,-,
   thick,base node/.style={circle,draw,minimum size=48pt}, real node/.style={double,circle,draw,minimum size=50pt},scale = 1.2]
   
   \node[shape=circle,draw=black](0) at(0,0) []{0.9};
   \node[shape=circle,draw=black](1) at(-3,-1) []{1};
   \node[shape=circle,draw=black](2) at(-1.5,-1) []{0};
   \node[shape=circle,draw=black](3) at(0,-1) []{0};
   \node[shape=circle,draw=black](4) at(1.5,-1) []{0};
   \node[shape=circle,draw=black](5) at(3,-1) []{0};

   \path[]
   (0) edge [ ]node {} (1)
   (0) edge [ ]node {} (2)
   (0) edge [ ]node {} (3)
   (0) edge [ ]node {} (4)
   (0) edge [ ]node {} (5);
      
\end{tikzpicture} 

 \caption{An example on k-tree.}
 \label{fig: an example on k-tree}
 \end{figure}
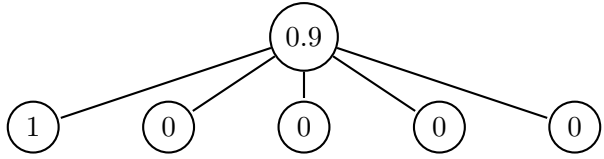

  The current greedy construction of the exploration set has limitations, 
  as is common with greedy policies. 
  Let us consider a scenario where the feedback graph is a k-tree, 
  as shown in \cref{fig: an example on k-tree}. 
  The node values represent rewards, and we assume the best arm is not the central node. 
  With the greedy construction, all nodes except the central one would be selected. 
  However, if the central node in the k-tree happens to have high rewards, 
  there is an opportunity to explore the other nodes while still achieving low regrets. 
  This trade-off is illustrated in the linear program \eqref{eq: sampling probability}.

 To enhance the sampling probabilities, we introduce a baseline probability $\tilde{p}_t(a)$, defined as:
 $$
 \tilde{p}_t(a) = \left\{
 \begin{array}{ll}
 \frac{1}{|S_t|+\gamma_t (\hat{f}_m(x_t,\hat{a}_t) - \hat{f}_m(x_t,a) )}, & \text{for all } a\in S_t-{ \hat{a}_t } \\
 0, & \text{for all } a\in \mathcal{A} - S_t \\
 1- \sum_{a\neq \hat{a}_t } \tilde{p}_t(a), & \text{for } a = \hat{a}_t,
 \end{array}
 \right.
 $$
 where $\hat{a}_t =\max_{a\in\mathcal{A}} \hat{f}_m(x_t,a)   $.

However, this approach may overlook highly informative actions that could provide valuable information when explored. 
To address this, we further improve the sampling probability by solving the following linear program:
\begin{equation}
\label{eq: sampling probability}
\begin{array}{lcl}
    & & \min_{p_t}  \sum_{a\in\mathcal{A}} p_t(a) (\hat{f}_m(x_t,\hat{a}_t)   -  \hat{f}_m(x_t,a) ) \\
    & s.t. & \sum_{a \in\mathcal{N}^{out}_{j}(G_t)} p_t(j) \geq \max_{a \in\mathcal{N}^{out}_{j}(G_t) } \tilde{p}_t(j), \forall a \neq \hat{a}_t \\
    &  & p_t(a) \geq 0, \forall a\in \mathcal{A} \\
    &  & \sum_{a \in \mathcal{A}} p_t(a) = 1.
  \end{array}
\end{equation}
The objective of this linear program is to minimize the expected one-step empirical regret with the baseline greedy policy $\pi_{\hat{f}_m}$. 
By optimizing the allocation of observation probabilities, this approach provides improved regret bounds, 
especially in scenarios with large gaps in rewards or the presence of highly informative actions.
The baseline observation probability guarantees minimax optimality, 
ensuring that the worst performance of our algorithm incurs at most $\sqrt{\alpha T \log |\mathcal{F}|}$ regret. 
The linear program efficiently allocates observation probabilities, 
leading to improved regret bounds in easy bandit instances where there are large gaps in rewards or the existence of highly informative actions.

\begin{myrem}
  An alternative approach to enhance the sampling probabilities is to consider the original inverse gap weighting sampling probability as the base sampling probability:
$$
\tilde{p}_t(a) = \left\{
\begin{array}{lcl}
    \frac{1}{|\mathcal{A}|+\gamma_m (\hat{f}_m(x_t,\hat{a}_t)   -  \hat{f}_m(x_t,a) )},  \text{ for all } a\neq \hat{a}_t  \\
    1- \sum_{a\neq \hat{a}_t }  \tilde{p}_t(a), \text{ for } a = \hat{a}_t ,
\end{array}
\right.
$$
where $\gamma_m = \sqrt{  \frac{\eta |\mathcal{A}| \tau_{m-1}  }{  2 \log (   2\delta^{-1}|\mathcal{A}||\mathcal{F}| T^2   )  }     }    $
This modification can lead to a more efficient exploration strategy in easy instances. 
However, since it covers all actions rather than just the selected ones, 
the minimax regret order will scale with $\sqrt{|\mathcal{A}| T \log |\mathcal{F}|}$, 
indicating worse performance on difficult instances compared to the baseline sampling probability in \eqref{eq: sampling probability}.
\end{myrem}

\textbf{Adaptive to gaps.}
Our algorithm incorporates adaptive features that effectively handle reward gaps. These features contribute to the overall performance improvement of \cref{alg: ADA-G}.
Firstly, the greedy construction of exploration set balance the trade-off between gaps and minimax optimality.
By putting arms with small gaps in priority,
we may improve the numerical performance of \cref{alg: ADA-G}.
Secondly, the linear program explores arms with the empirically minimal regrets, 
resulting in potentially low regrets despite exploring numerous arms.
By allocating observation probabilities based on the minimal regrets, the linear program effectively samples actions that provide valuable information, contributing to improved regret bounds.
Thirdly, to derive gap-dependent upper bounds in contextual bandit problems, we introduce an instance-dependent scale factor $\lambda_m$ into the parameter $\gamma_t$. 
This scale factor provides a sample-based approximation to the quantity:
$
\mathbb{P}_{\mathcal{X}} ( |\mathcal{A}(x;\mathcal{F}_m) | > 1 ) / \sqrt{  \mathbb{P}_{\mathcal{X}} ( |\mathcal{A}(x;\mathcal{F}_{m-1}) | > 1 )     }.
$

It has been pointed out by \cite{instanceCB_RL} that obtaining a gap-dependent regret for general contextual bandits is not possible, even under the uniform gap assumption.
To overcome this limitation, researchers have developed algorithms that achieve instance-dependent bounds for specific classes of problems under additional structural or distributional assumptions.
Similar to \cite{instanceCB_RL}, we consider the following policy disagreement coefficient 
\begin{equation}
  \label{eq: policy disagreement coefficient}
\begin{aligned}
&\theta^{pol}(\mathcal{F},\epsilon_0) \\
=& \sup_{\epsilon \geq \epsilon_0} \frac{1}{\epsilon} \mathbb{P}_{\mathcal{X}}(x \in \mathcal{X}: \exists f \in \mathcal{F}_{\epsilon} \text{ such that } \pi_f(x)\neq \pi_{f^*}(x)   ),
\end{aligned}
\end{equation}
where $\mathcal{F}_\epsilon = \{f\in\mathcal{F}| \mathbb{P}_{\mathcal{X}}(x \in \mathcal{X}: \pi_f(x) \neq \pi_{f^*}(x)    ) \leq \epsilon \}$.
It can be observed that $\mathbb{P}_{\mathcal{X}} ( |\mathcal{A}(x;\mathcal{F}_m) | > 1 ) \leq \mathbb{P}_{\mathcal{X}}(x \in \mathcal{X}: \exists f \in \mathcal{F}_{m} \text{ such that } \pi_f(x)\neq \pi_{f^*}(x)   ) $, 
so the disagreement coefficient can provide a valid coefficient in upper bounds of the instance-dependent regrets.


\subsection{Regret analysis} 
Our regret analysis builds on a framework established in \cite{fasterCB,instanceCB_RL}, 
which analyzes contextual bandit algorithms in the universal policy space $\Psi:= \prod_{x\in\mathcal{X}} \mathcal{A}$. 

Suppose that the high-probability event $\Gamma$ in \cref{lem: high-probability events} holds.
Then we have $\hat{f}_m \in \mathcal{F}_m$ and $f^* \in \mathcal{F}_m$ for all epoch $m$.
To analyze regrets, for all rounds $t$ in the epoch $m$, it suffices to consider the followings:
$$
\mathcal{R}^{dis}_t(\pi) =  \mathbb{E}_{x}[ \ind{ |\mathcal{A}(x;\mathcal{F}_m)|>1  }   f^*(x,\pi(x))]  
$$
and 
$$
Reg_t(\pi) = \mathcal{R}_t^{dis}(\pi_{f^*}) -\mathcal{R}_t^{dis}(\pi).
$$
Furthermore, we use the empirical best policy $\pi_{\hat{f}_{m}}$ as the baseline and define
$$
\widehat{\mathcal{R}}_t(\pi) = \mathbb{E}_{x_t}[\ind{ |\mathcal{A}(x;\mathcal{F}_m)|>1  }  \hat{f}_{m}(x_t,\pi(x_t))]
$$
and   
$$ 
\widehat{ Reg}_t(\pi) =   \mathbb{E}_{x}[\widehat{\mathcal{R}}_t(\pi_{\hat{f}_{m}}) - \widehat{\mathcal{R}}_t(\pi)].
$$

For any realization $S_t$, $\gamma_t$ and $\hat{f}_m$, let $Q_t(\cdot)$ be the equivalent policy distribution for $p_t(\cdot | \cdot)$, i.e.,
$$
Q_t(\pi) = \prod_{x\in\mathcal{X}}   p_t(\pi(x)|x) , \forall \pi\in\Psi.
$$
The existence and uniqueness of such measure $Q_t(\cdot)$ is a corollary of Kolmogorov's extension theorem.
Note that both $\Psi$ and $Q_t(\cdot)$ are $\mathcal{H}_{t-1}$-measurable, 
where $\mathcal{H}_{t-1}$ is the filtration up to the time $t-1$.
We refer to Section 3.2 of \cite{fasterCB} for more detailed intuition for $Q_t(\cdot)$ and proof of existence. 
By Lemma 4 of \cite{fasterCB}, we know that for all epoch $m$ and all rounds $t$ in epoch $m$,
we can rewrite the expected regret in terms of our notations as 
$$
\mean{Reg(T)} = \sum_{t=1}^T \sum_{ \pi \in \Psi} Q_{t}(\pi) Reg(\pi).
$$
For simplicity, we define an epoch-dependent quantities 
$$
\rho_1 = 1, \rho_m =  \sqrt{  \frac{\eta  (\tau_{m-1} -\tau_{m-2}) }{  2 \log (   2\delta^{-1}|\mathcal{A}||\mathcal{F}| T^2   )  }     }, m\geq 2,
$$
so $\gamma_t = \sqrt{|S_t|} \rho_{m(t)}   $ for $m(t)\geq 2$.

In the graph feedback setting, a modification is required to decouple the action sampling probability and the observation probability, which differs from the original inverse gap weighting technique.
The observation probability of certain action is defined as $q_t(a) = \sum_{a \in\mathcal{N}^{out}_{j}(G_t)} p_t(j)$. 
This construction gives rise to the following implicit optimization problem.



\begin{mylem}
  \label{lem: IOP}
(Implicit Optimization Problem). For all epoch $m$ and all rounds $t$ in epoch $m$, $Q_t$ is a feasible solution to the following 
implicit optimization problem:
\begin{align*}
  & \sum_{\pi\in \Psi }  Q_t(\pi) \widehat{ Reg}_m(\pi) \leq  \sqrt{ |S_t| -1 } / \rho_m,   \\
  &  \mathbb{E}_{x}\left[ \frac{ 1 }{q_t(\pi(x)|x)} \right]  \leq  |S_t|  +    \sqrt{ |S_t| } \rho_m    \widehat{ Reg}_m(\pi), \forall \pi \in \Psi.
\end{align*}
\end{mylem}
We further refine this lemma to \cref{lem: disagreement-based IOP},
which extends the original IOP from \cite{fasterCB,instanceCB_RL} to the graph feedback setting.
In \cref{lem: disagreement-based IOP}, we introduce the quantity $q_m = \mathbb{P}_{\mathcal{X}}(|\mathcal{A}(x;\mathcal{F}_m)| > 1)$, 
which allows us to establish a connection between regret and the policy disagreement coefficient \eqref{eq: policy disagreement coefficient}.

Intuitively, the empirically greedy policy should approach the optimal policy (the first constraint). 
However, to ensure optimality, the algorithm also need to explore around this greedy policy (the second constraint).
Such trade-off leads to \cref{lem: IOP} and the graph feedback setting makes the RHS scale with the size of exploration sets rather than the whole action set.

Since the greedy policy $\pi_{\hat{f}_m}$ serves as a relatively good baseline, 
we can bound the true regret as the sum of the regret with respect to $\pi_{\hat{f}_m}$ and the distance between $\pi_{\hat{f}_m}$ and $\pi_{f^*}$. 
Detailed results are provided in \cref{lem: prediction error} and \cref{lem: prediction error of regrets of gap-dependent upper bound}. 
These lemmas collectively lead to the derivation of a gap-dependent upper bound.

\begin{mythm}
  \label{thm: gap-dependent upper bound}
  Consider a contextual bandit problem with graph feedbacks. 
  Suppose that \cref{asp: realizability} holds.
  Then for any instance with uniform gap $\Delta>0$, 
  with probability at least $1-\delta$,
  the expected regret $\mathbb{E}[{Reg(T)}] $ of \cref{alg: ADA-G} is upper bounded by 
  $$
  \mathcal{O}\left( \min_{\epsilon>0} \max\Big\{ \epsilon\Delta T,    \frac{ \theta^{pol}(\mathcal{F},\epsilon)  \mathbb{E}_{G}[ \alpha(G)  ]  \log(\delta^{-1}T^2|\mathcal{F}|) \log T    }{ \Delta  }        \Big\}   \right).
  $$
  Here, the expectation in regret is taken with respect to the randomness in the algorithm. 
\end{mythm}

\textbf{Minimax optimality.}
To show the optimality of our algorithms, we prove the following minimax regret upper bounds.

\begin{mythm}
  \label{thm: minimax upper bound of FALCON.G}
  Consider a contextual bandit problem with graph feedbacks. 
  Suppose that the realizability \cref{asp: realizability} holds.
  Then with probability at least $1-\delta$, the expected regret $\mathbb{E}[{Reg(T)}] $ of \cref{alg: ADA-G} is upper bounded by 
  $\mathcal{O} \left(\sqrt{ \mathbb{E}_{G}[ \alpha(G)  ] T \log(\delta^{-1}|\mathcal{F}| \log T   \log (|\mathcal{A}|T)   ) } \right)$.
Here, the expectation of regrets is taken with respect to the randomness of the algorithm.
\end{mythm}

Several previous studies, such as \cite{MAB-GB,beyondBandits}, 
have established an information-theoretic lower bound on the minimax regret for adversarial graph bandit problems. 
The fundamental idea behind these bounds is to select a maximal independence set and transform the original $|\mathcal{A}|$-armed bandit learning problem into $\alpha$-armed bandit problems.
By exploiting the properties of the independence set, the rewards of arms outside the selected set are effectively set to zero. 
Consequently, no algorithm can achieve a performance better than $\sqrt{\alpha T}$.
In the contextual setting, we can make reduce this new setting to original contextual bandit with $\alpha$ actions.
Hence, according to \cite{CBwithPredictableRewards}, the lower bound will be $\sqrt{\alpha T \log |\mathcal{F}|} $.

 Our \cref{alg: ADA-G} can be applicable to multi-armed bandit (MAB) instances, 
 and it retains the validity of \cref{thm: gap-dependent upper bound}. 
 Notably, in comparison to the MAB algorithm proposed in \cite{MAB-GB}, 
 our approach achieves regrets of order ${\mathcal{O}}\left(\sqrt{\mathbb{E}_G[\alpha(G)] T \log(\delta^{-1}|\mathcal{F}|\log^2 (|\mathcal{A}|T)  )} \right)$, 
 with an additional factor $\log|\mathcal{F}|$.
 The extra $\log|\mathcal{F}|$ factor comes from the complexity of contextual information, 
 which is unavoidable in our setting.



%% file: numerical.tex
\section{Numerical Experiments}
\label{sec: numerical experiments}

\textbf{Experimental setup.}
In numerical experiments, we randomly generate a function space $\mathcal{F}=\{(x-x_0)^\top (a-a_0) \}$ with a size of $50$ by sampling $x_0$ and $a_0$ in $\mathbb{R}^d$ from standard normal distributions, where $d=10$.
We then randomly choose a function as the true reward function $f^*$ from $\mathcal{F}$, 
and generate the reward as
$
Y= f^*(X,A) + \mathcal{N}(0,1),
$
where the context $X$ is drawn i.d.d. from $\mathcal{N}(0,1)$ and $A$ is the selected action. 
The whole action set $\mathcal{A}$ is randomly initialized from $[-1,1]^d$.
We repeat each instance $40$ times to obtain a smooth regret curve.
The hyperparameters in \cref{alg: ADA-G} are set to be $\eta=1$ and $\delta = 0.1$.

\textbf{Ablation experiments.}
In this part, we use empirical results to demonstrate the significant benefits of \cref{alg: ADA-G}
in leveraging the graph feedback structure. 
To verify its effectiveness, we conduct numerical ablation experiments in time-varying graphs with different types. 
Our baseline is FALCON \cite{fasterCB}, with additional observations only used for estimating $\hat{f}_m$ at the epoch $m$.
It is important to note that the method presented in \cite{contextualGB} relies on an online oracle and a fixed global hyperparameter $\gamma$, 
making it challenging to satisfy these conditions in practical scenarios. 
To address this limitation, we modify their method so that the offline oracle is applicable, 
providing a practical and effective alternative to the approach in \cite{contextualGB}.

\begin{figure*}[htbp]

    \centering
\begin{tikzpicture}
    \begin{customlegend}[legend columns=4,legend style={draw=none,column sep=2ex,nodes={scale=0.6, transform shape}},
        legend entries={
                        \text{\textbf{FALCON} \cite{fasterCB}},
                        \text{modified \textbf{RegCB.G} \cite{contextualGB}},
                        \text{ours}
                        }]

        \addlegendimage{color=red}
        \addlegendimage{color=blue}
        \addlegendimage{color=green}
    \end{customlegend}
\end{tikzpicture}

\begin{tabular}{lcr}

\begin{tikzpicture}[scale=0.8]
    \begin{axis}[
        height = 0.4\textwidth,
        width = 0.38\textwidth,
        xlabel = time $t$,   
        ylabel = regret,
        ymin=0,
        ymax=1600,
        xmin=0,
        xtick pos = left,
        ytick pos = left,
    ]

    \addplot [color=red, thick ] table [x index=0,y index=1, col sep = comma] {data/org_mean_regret_gtype_Ktree_repeat_50_K_50.csv};
    \addplot [color=blue, thick ] table [x index=0,y index=1, col sep = comma] {data/ind_mean_regret_gtype_Ktree_repeat_50_K_50.csv};
    \addplot [color=green, thick ] table [x index=0,y index=1, col sep = comma] {data/ada_mean_regret_gtype_Ktree_repeat_50_K_50.csv};

    \addplot [name path=option1_upper, color=red, dashed ] table [x index=0,y index=1, col sep = comma] {data/org_upper_regret_gtype_Ktree_repeat_50_K_50.csv};
    \addplot [name path=option1_down, color=red, dashed ] table [x index=0,y index=1, col sep = comma] {data/org_lower_regret_gtype_Ktree_repeat_50_K_50.csv};

    \addplot [name path=option2_upper, color=blue, dashed ] table [x index=0,y index=1, col sep = comma] {data/ind_upper_regret_gtype_Ktree_repeat_50_K_50.csv};
    \addplot [name path=option2_down, color=blue, dashed ] table [x index=0,y index=1, col sep = comma] {data/ind_lower_regret_gtype_Ktree_repeat_50_K_50.csv};

    \addplot [name path=option3_upper,color=green, dashed ] table [x index=0,y index=1, col sep = comma] {data/ada_upper_regret_gtype_Ktree_repeat_50_K_50.csv};
    \addplot [name path=option3_down, color=green, dashed ] table [x index=0,y index=1, col sep = comma] {data/ada_lower_regret_gtype_Ktree_repeat_50_K_50.csv};
    
    \addplot [red!50,fill opacity=0.5] fill between[of=option1_upper and option1_down];  
    \addplot [blue!50,fill opacity=0.5] fill between[of=option2_upper and option2_down];  
    \addplot [green!50,fill opacity=0.5] fill between[of=option3_upper and option3_down];  
  
    \end{axis}
\end{tikzpicture}
&
\begin{tikzpicture}[scale=0.8]
    \begin{axis}[
        height = 0.4\textwidth,
        width = 0.38\textwidth,
        xlabel = time $t$,   
        ymax=1600,
        ymin=0,
        xmin=0,
        xtick pos = left,
        ytick pos = left,
    ]

    \addplot [color=red, thick ] table [x index=0,y index=1, col sep = comma] {data/org_mean_regret_gtype_clique_group_repeat_50_K_50.csv};
    \addplot [color=blue, thick ] table [x index=0,y index=1, col sep = comma] {data/ind_mean_regret_gtype_clique_group_repeat_50_K_50.csv};
    \addplot [color=green, thick ] table [x index=0,y index=1, col sep = comma] {data/ada_mean_regret_gtype_clique_group_repeat_50_K_50.csv};

    \addplot [name path=option1_upper, color=red, dashed ] table [x index=0,y index=1, col sep = comma] {data/org_upper_regret_gtype_clique_group_repeat_50_K_50.csv};
    \addplot [name path=option1_down, color=red, dashed ] table [x index=0,y index=1, col sep = comma] {data/org_lower_regret_gtype_clique_group_repeat_50_K_50.csv};

    \addplot [name path=option2_upper, color=blue, dashed ] table [x index=0,y index=1, col sep = comma] {data/ind_upper_regret_gtype_clique_group_repeat_50_K_50.csv};
    \addplot [name path=option2_down, color=blue, dashed ] table [x index=0,y index=1, col sep = comma] {data/ind_lower_regret_gtype_clique_group_repeat_50_K_50.csv};

    \addplot [name path=option3_upper,color=green, dashed ] table [x index=0,y index=1, col sep = comma] {data/ada_upper_regret_gtype_clique_group_repeat_50_K_50.csv};
    \addplot [name path=option3_down, color=green, dashed ] table [x index=0,y index=1, col sep = comma] {data/ada_lower_regret_gtype_clique_group_repeat_50_K_50.csv};
    
    \addplot [red!50,fill opacity=0.5] fill between[of=option1_upper and option1_down];  
    \addplot [blue!50,fill opacity=0.5] fill between[of=option2_upper and option2_down];  
    \addplot [green!50,fill opacity=0.5] fill between[of=option3_upper and option3_down];  

    \end{axis}
\end{tikzpicture}
&

\begin{tikzpicture}[scale=0.8]
    \begin{axis}[
        height = 0.4\textwidth,
        width = 0.38\textwidth,
        xlabel = time $t$,   
        ymax=1600,
        ymin=0,
        xmin=0,
        xtick pos = left,
        ytick pos = left,
    ]

    \addplot [color=red, thick ] table [x index=0,y index=1, col sep = comma] {data/org_mean_regret_gtype_random_repeat_50_K_50.csv};
    \addplot [color=blue, thick ] table [x index=0,y index=1, col sep = comma] {data/ind_mean_regret_gtype_random_repeat_50_K_50.csv};
    \addplot [color=green, thick ] table [x index=0,y index=1, col sep = comma] {data/ada_mean_regret_gtype_random_repeat_50_K_50.csv};

    \addplot [name path=option1_upper, color=red, dashed ] table [x index=0,y index=1, col sep = comma] {data/org_upper_regret_gtype_random_repeat_50_K_50.csv};
    \addplot [name path=option1_down, color=red, dashed ] table [x index=0,y index=1, col sep = comma] {data/org_lower_regret_gtype_random_repeat_50_K_50.csv};

    \addplot [name path=option2_upper, color=blue, dashed ] table [x index=0,y index=1, col sep = comma] {data/ind_upper_regret_gtype_random_repeat_50_K_50.csv};
    \addplot [name path=option2_down, color=blue, dashed ] table [x index=0,y index=1, col sep = comma] {data/ind_lower_regret_gtype_random_repeat_50_K_50.csv};

    \addplot [name path=option3_upper,color=green, dashed ] table [x index=0,y index=1, col sep = comma] {data/ada_upper_regret_gtype_random_repeat_50_K_50.csv};
    \addplot [name path=option3_down, color=green, dashed ] table [x index=0,y index=1, col sep = comma] {data/ada_lower_regret_gtype_random_repeat_50_K_50.csv};
    
    \addplot [red!50,fill opacity=0.5] fill between[of=option1_upper and option1_down];  
    \addplot [blue!50,fill opacity=0.5] fill between[of=option2_upper and option2_down];  
    \addplot [green!50,fill opacity=0.5] fill between[of=option3_upper and option3_down];

    \end{axis}
\end{tikzpicture}

\end{tabular}
\caption{Comparison with baselines and our algorithms in graph feedback setting. We conduct numerical experiments on k-tree, clique-group and random graphs, respectively.
The top and bottom dashed curves represent the curves obtained by adding and subtracting one standard deviation to the regret curve of the corresponding color, respectively.}
\label{fig: comparison with classical and our algorithms in graph feedback settings}
\end{figure*}
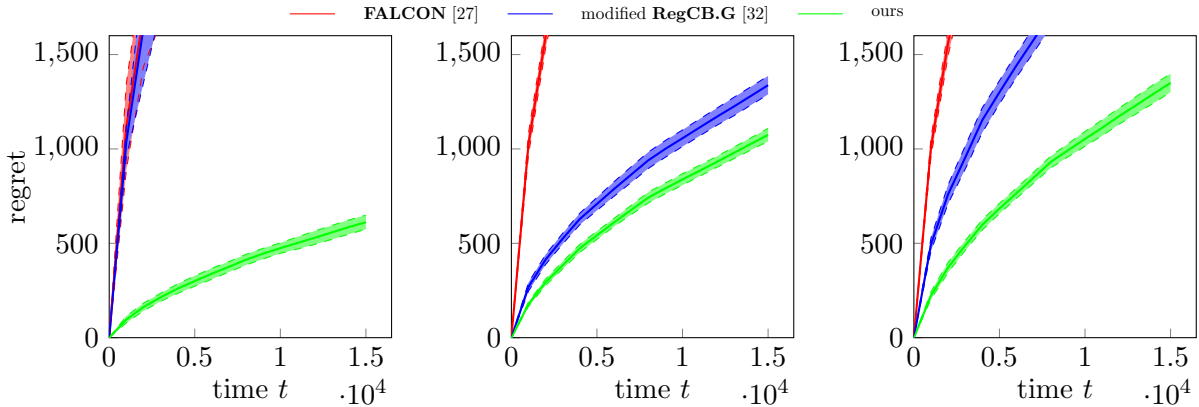

In our ablation experiments, we generate time-varying graphs to test performance of our algorithm on different graph types.
The results depicted in \cref{fig: comparison with classical and our algorithms in graph feedback settings} clearly demonstrate 
that the average regrets of \cref{alg: ADA-G} are significantly smaller than those of \cite{fasterCB} and \cite{contextualGB}.

We begin by conducting numerical experiments on k-trees, 
which are star-shaped structures commonly seen in social networks.
These graphs represent scenarios where there exist influential individuals within the network. 
Specifically, the independence number of such graphs scales with the size of action set.
During our experiments, we observe that the method proposed in \cite{contextualGB} slightly improves the regret bound compared to \cite{fasterCB}. 
However, when comparing their methods to ours, there is a significant gap between the resulting regret curves. 
Our approach outperforms both \cite{fasterCB} and \cite{contextualGB} in terms of regret, 
demonstrating its effectiveness on k-tree graphs.

Next, we perform experiments on clique-groups, 
which are graphs specifically designed with each block consisting of a perfect graph. 
This construction ensures that the independence number of the graph is equal to the clique-partition number, 
making it suitable for modeling communities within social networks.
In this case, the greedy construction of $\tilde{p}_t$ is sufficient in achieving relatively good performance. 
Therefore, the improvement achieved by our method is limited, compared with that in \cite{contextualGB}.
Nonetheless, we still observe a slight improvement, which can be attributed to the adaptability of our approach to instance gaps.

Lastly, we conduct numerical experiments on time-varying random graphs, 
which can be considered as a combination of the previously mentioned graph types. 
We observe that the performance of \cite{fasterCB} does not vary significantly across different types of graphs. 
This finding suggests that relying solely on additional observations to estimate $\hat{f}$ is insufficient in capturing the underlying dynamics of the graph.
In contrast, our \cref{alg: ADA-G} proposes a specialized design for action sampling probabilities in graph feedback settings, 
leading to improved performance compared to the baselines. 
Notably, our algorithm does not rely on difficult-to-obtain graph parameters, 
further enhancing its practicality. 
These results underscore the adaptive capability of our approach when dealing with random graphs, 
demonstrating its effectiveness in capturing and leveraging the dynamics and structure of the graph to enhance decision-making.

\textbf{Test on graph parameters.}
We test our algorithm in clique-group (as shown in \cref{fig: different types of graphs}), 
where the independence number $\alpha(G)$ is equal to the number of cliques.
To showcase that our method does not scale with the size of the action set size,
we report the average cumulative regrets over $2^{11}$ rounds and their standard deviation in \cref{tb: test on clique-group}.
Additionally, the regrets curves of different action set size are depicted in \cref{fig: regret curves on clique groups}.

Analyzing the results in \cref{tb: test on clique-group}, we observe that the cumulative regrets slightly decreases as the number of arms increases. 
This indicates that the regrets do not scale with $|\mathcal{A}|$ since the independence number remains fixed for each learning process.
This numerical result validates the findings of our \cref{thm: minimax upper bound of FALCON.G}.
Furthermore, the decreasing trend can be attributed to the fact that when the learner explores suboptimal arms, 
the rewards of these suboptimal arms are more likely to be high when the number of arms is larger. 
Hence, this serves as an evidence of the capability of our algorithm to adapt to instance gaps.

\begin{table}[h]
    \centering

    \begin{tabular}{|c|c|c|c|c|c|}
        \hline
        size of the action set                                                       & 20                                                       & 40                                                       & 60                                                       & 80                                                       & 100                                                      \\ \hline
        \begin{tabular}[c]{@{}c@{}}mean(std) of the\\ cumulative regret\end{tabular} & \begin{tabular}[c]{@{}c@{}}329.86\\ (31.18)\end{tabular} & \begin{tabular}[c]{@{}c@{}}311.34\\ (27.21)\end{tabular} & \begin{tabular}[c]{@{}c@{}}300.64\\ (24.11)\end{tabular} & \begin{tabular}[c]{@{}c@{}}292.60\\ (29.59)\end{tabular} & \begin{tabular}[c]{@{}c@{}}277.35\\ (23.00)\end{tabular} \\ \hline
    \end{tabular}
    
    \caption{Test on clique-group with its clique number equal to 5.}
    \label{tb: test on clique-group}
\end{table}

\textbf{Test on real-world dataset.}
We conduct an empirical evaluation of our algorithm using the dataset Flixster. 
This is a social network dataset crawled by \cite{Flixster}, 
and consists of $1$ million users and $14$ million friendship relations.
To make the experiments more manageable, we employ the union-find algorithm to find a strongly connected subgraph with a size of $100$. 
Constructing such subgraphs from the entire graph can be time-consuming due to its large scale. 
Therefore, we generate a collection of 100 random subgraphs in advance and store them for further experiments.
At each round, we randomly sample such a subgraph from the graph collection as a feedback graph at this round.

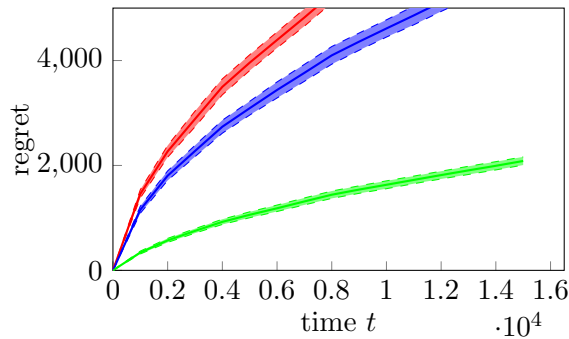
\begin{figure}[ht]

    \centering
     \begin{tikzpicture}[scale=0.8]
       \begin{axis}[
           height = 0.36\textwidth,
           width = 0.55\textwidth,
           xlabel = time $t$,   
           ylabel = regret,
           ymax = 5000,
           ymin=0,
           xmin=0,
           xtick pos = left,
           ytick pos = left,
           legend style={at={(0.1,0.75)}, anchor=west, nodes={scale=0.8, transform shape}}
       ]

       \addplot [color=red, thick ] table [x index=0,y index=1, col sep = comma] {data/org_mean_regret_gtype_Flixster_repeat_50_K_100.csv};
       \addplot [color=blue, thick ] table [x index=0,y index=1, col sep = comma] {data/ind_mean_regret_gtype_Flixster_repeat_50_K_100.csv};
       \addplot [color=green, thick ] table [x index=0,y index=1, col sep = comma] {data/ada_mean_regret_gtype_Flixster_repeat_50_K_100.csv};

       \addplot [name path=org_upper, color=red, dashed ] table [x index=0,y index=1, col sep = comma] {data/org_upper_regret_gtype_Flixster_repeat_50_K_100.csv};
       \addplot [name path=org_lower, color=red, dashed ] table [x index=0,y index=1, col sep = comma] {data/org_lower_regret_gtype_Flixster_repeat_50_K_100.csv};
       \addplot [red!50,fill opacity=0.5] fill between[of=org_upper and org_lower]; 
 
       \addplot [name path=ind_upper, color=blue, dashed ] table [x index=0,y index=1, col sep = comma] {data/ind_upper_regret_gtype_Flixster_repeat_50_K_100.csv};
       \addplot [name path=ind_lower, color=blue, dashed ] table [x index=0,y index=1, col sep = comma] {data/ind_lower_regret_gtype_Flixster_repeat_50_K_100.csv};
       \addplot [blue!50,fill opacity=0.5] fill between[of=ind_upper and ind_lower]; 

       \addplot [name path=ada_upper, color=green, dashed ] table [x index=0,y index=1, col sep = comma] {data/ada_upper_regret_gtype_Flixster_repeat_50_K_100.csv};
       \addplot [name path=ada_lower, color=green, dashed ] table [x index=0,y index=1, col sep = comma] {data/ada_lower_regret_gtype_Flixster_repeat_50_K_100.csv};
       \addplot [green!50,fill opacity=0.5] fill between[of=ada_upper and ada_lower];

       \end{axis}
    \end{tikzpicture}
    \caption{Regret curves on Flixster social network.}
    \label{fig: regret curves on real-world dataset}
    \end{figure}

Real-world social networks often exhibit a combination of star-shaped structures and dense cliques. This characteristic is reflected in the dataset we used from the Flixster social network. 
In \cref{fig: regret curves on real-world dataset}, we can observe that our method continues to demonstrate advantages over the baselines when applied to this network.
As we discussed in the ablation experiments, the improvement in performance can be attributed to two factors: the underlying graph structures and the reward gaps.
The numerical results obtained from the real-world dataset further validate the effectiveness of our algorithm in practical applications. 
It demonstrates that our approach is not only applicable to synthetic graphs but also performs well on real-world networks, 
capturing their unique characteristics and optimizing decision-making processes accordingly.

%% file: conclusion.tex
\section{Conclusion}

In this paper, we have introduced a framework for incorporating side-observations into contextual bandits with a general reward function space. 
We have derived instance-independent upper bounds on the regret and proposed a near-optimal algorithm that matches corresponding lower bounds up to logarithmic terms and constants. 
However, there are several avenues for future research and extension of our work.

Firstly, it would be valuable to explore the gap-dependent upper bound in a more precise manner than what is presented in \cref{thm: gap-dependent upper bound}. 
Assuming the gap condition:
$
f^*(x,\pi_{f^*}(x)) - f^*(x,a) \geq \Delta_a, \forall x\in\mathcal{X},
$
we still lack a method to establish a gap-dependent upper bound specific to each arm. 
Obtaining such gap-dependent upper bounds would allow us to more accurately balance the trade-off between the number of arms to explore and their corresponding gaps.


Additionally, it is still unresolved to derive gap-dependent upper bounds for more general types of feedback graphs, 
such as weakly observable graphs. 
Our current approach may face limitations in constructing an effective baseline probability in these cases.
By extending our framework to handle diverse types of feedback graphs, 
we can enhance its versatility and address a wider range of real-world applications.

%% file: appendix.tex
 \section{Additional Numerical Experiments}

In \cref{sec: numerical experiments}, we implement algorithms in various graphs to verify our findings. 
These graphs are created using our random graph generator, which is described in \cref{alg: random graph generator}. 
To simplify the following graphs, self-loops are omitted. 
We present these graphs in \cref{fig: different types of graphs}.

\begin{figure}[h]
    \center
 \begin{tabular}{ccccc}
       \begin{tikzpicture}[shorten >=1pt,auto,node distance=4cm,-,
             thick,base node/.style={circle,draw,minimum size=48pt}, real node/.style={double,circle,draw,minimum size=50pt},scale = 1.2]
             
             \node[shape=circle,draw=black](0) at(0,0) []{};
             \node[shape=circle,draw=black](1) at(0,-1) []{};
             \node[shape=circle,draw=black](2) at(1,0) []{};
             \node[shape=circle,draw=black](3) at(1,-1) []{};
     
             \path[]
             (0) edge []node {} (1)
             (0) edge []node {} (3)
             (2) edge []node {} (1)
             (2) edge []node {} (3)
             (1) edge []node {} (3)
             (2) edge []node {} (0);
                
       \end{tikzpicture}

    &
    \begin{tikzpicture}[shorten >=1pt,auto,node distance=1cm,-,
      thick,base node/.style={circle,draw,minimum size=48pt}, real node/.style={double,circle,draw,minimum size=50pt},scale = 1.2]
      
      \node[shape=circle,draw=black](0) at(0,0) []{};
      \node[shape=circle,draw=black](1) at(0.5,0) []{};
      \node[shape=circle,draw=black](2) at(1,0) []{};
      \node[shape=circle,draw=black](3) at(1.5,0) []{};
      \node[shape=circle,draw=black](4) at(2,0) []{};
      \node[shape=circle,draw=black](5) at(2.5,0) []{};
      \node[shape=circle,draw=black](6) at(0.25,0.5) []{};
      \node[shape=circle,draw=black](7) at(1.25,0.5) []{};
      \node[shape=circle,draw=black](8) at(2.25,0.5) []{};

      \path[]
      (0) edge []node {} (1)
      (1) edge []node {} (2)
      (2) edge []node {} (3)
      (3) edge []node {} (4)
      (4) edge []node {} (5)
      (6) edge []node {} (0)
      (6) edge []node {} (1)
      (7) edge []node {} (2)
      (7) edge []node {} (3)
      (8) edge []node {} (4)
      (8) edge []node {} (5);        
\end{tikzpicture}
&

\begin{tikzpicture}[shorten >=1pt,auto,node distance=4cm,-,
   thick,base node/.style={circle,draw,minimum size=48pt}, real node/.style={double,circle,draw,minimum size=50pt},scale = 1.2]
   
   \node[shape=circle,draw=black](0) at(0,0) []{};
   \node[shape=circle,draw=black](1) at(-1,-1) []{};
   \node[shape=circle,draw=black](2) at(-0.5,-1) []{};
   \node[shape=circle,draw=black](3) at(0,-1) []{};
   \node[shape=circle,draw=black](4) at(0.5,-1) []{};
   \node[shape=circle,draw=black](5) at(1,-1) []{};

   \path[]
   (0) edge [ ]node {} (1)
   (0) edge [ ]node {} (2)
   (0) edge [ ]node {} (3)
   (0) edge [ ]node {} (4)
   (0) edge [ ]node {} (5);
      
\end{tikzpicture} 
&
 
  \begin{tikzpicture}[shorten >=1pt,auto,node distance=1cm,-,
    thick,base node/.style={circle,draw,minimum size=48pt}, real node/.style={double,circle,draw,minimum size=50pt},scale = 1.2]
    
    \node[shape=circle,draw=black](0) at(0,0) []{};
    \node[shape=circle,draw=black](1) at(1,0) []{};
    \node[shape=circle,draw=black](2) at(0,1) []{};
    \node[shape=circle,draw=black](3) at(1,1) []{};
    \node[shape=circle,draw=black](4) at(-0.3,0.5) []{};
    \node[shape=circle,draw=black](5) at(1.5,0.4) []{};
 
    \path[]
    (0) edge []node {} (1)
    (1) edge []node {} (2)
    (4) edge []node {} (2)
    (5) edge []node {} (1)
    (5) edge []node {} (2)
    (3) edge []node {} (2)
    (5) edge []node {} (3);
 \end{tikzpicture}

 \end{tabular}
 \caption{Different types of graphs in order: a fully connected graph, a clique group, a k-tree, a random graph.}
 \label{fig: different types of graphs}
 \end{figure}
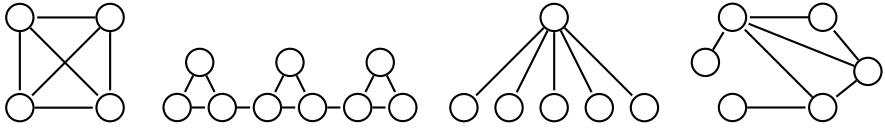

 \begin{figure}[ht]

   \centering
    \begin{tikzpicture}[scale=0.8]
      \begin{axis}[
          height = 0.45\textwidth,
          width = 0.55\textwidth,
          xlabel = time $t$,   
          ylabel = regret,
          ymax=400,
          ymin=0,
          xmin=0,
          xtick pos = left,
          ytick pos = left,
          legend style={at={(0.1,0.75)}, anchor=west, nodes={scale=0.8, transform shape}}
      ]

      \addplot [color=blue, thick ] table [x index=0,y index=1, col sep = comma] {data/ada_mean_regret_gtype_clique_group_repeat_40_K_20.csv};

      \addplot [color=red, thick ] table [x index=0,y index=1, col sep = comma] {data/ada_mean_regret_gtype_clique_group_repeat_40_K_40.csv};

      \addplot [color=yellow, thick ] table [x index=0,y index=1, col sep = comma] {data/ada_mean_regret_gtype_clique_group_repeat_40_K_60.csv};

      \addplot [color=green, thick ] table [x index=0,y index=1, col sep = comma] {data/ada_mean_regret_gtype_clique_group_repeat_40_K_80.csv};

      \addplot [color=black, thick ] table [x index=0,y index=1, col sep = comma] {data/ada_mean_regret_gtype_clique_group_repeat_40_K_100.csv};

      \legend{
         $|\mathcal{A}|=20$,
         $|\mathcal{A}|=40$,
         $|\mathcal{A}|=60$,
         $|\mathcal{A}|=80$,
         $|\mathcal{A}|=100$,
      }

      \end{axis}
   \end{tikzpicture}
   \caption{Regret curves on clique groups.}
   \label{fig: regret curves on clique groups}
   \end{figure}
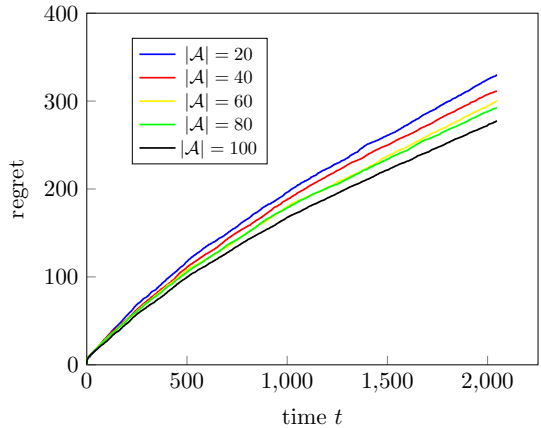

\section{Proofs}

\subsection{Notations}

To incorporate the instance gaps, we define the following quantities that are used to show the regret upper bound:
\begin{align*}
   & \nu_m = \mathbb{E}_{x}[|\mathcal{A}(x;\mathcal{F}_m)|>1] \\
   & \hat{\nu}_m = \mathbb{E}_{x \sim \mathcal{D}_m}[|\mathcal{A}(x;\mathcal{F}_m)|>1] \\
   & w_m = \nu_m + \mu_m \\
   & \hat{w}_m = \hat{\nu}_m + \mu_m.
\end{align*}
Similar to the framework of \cite{instanceCB_RL}, we need to define the following high-probability event $\Gamma$.

\begin{mylem}[\cite{CBwithOracle,instanceCB_RL}]
   \label{lem: high-probability events}
   Let $C_{\delta} = 16\log (   2\delta^{-1} |\mathcal{F}| |\mathcal{A}| T^2  )$ and $M = \lceil \log T  \rceil$. 
   Denote the following events as $\Gamma_i,i=1,2,3$, respectively.
   \begin{enumerate}
      \item For all $m \in [ M]$, for all $\beta_m\geq 0$, 
      $$
      \sum_{t = \tau_{m-2}+1}^{\tau_{m-1}} \mathbb{E}_{x_t,a_t}[ \sum_{a\in\mathcal{N}_{a_t}(G_t)} ( \hat{f}_m(x_t,a) -f^*(x_t,a)   )^2 |\mathcal{H}_{t-1}   ] \leq C_{\delta}
      $$
      and 
      $$
      \forall f\in\mathcal{F}_m, \sum_{t = \tau_{m-2}+1}^{\tau_{m-1}} \mathbb{E}_{x_t,a_t}[\sum_{a\in\mathcal{N}_{a_t}(G_t)} ( \hat{f}_m(x_t,a) -f^*(x_t,a)   )^2 |\mathcal{H}_{t-1}   ] \leq 2\beta_m+C_{\delta}.
      $$
      \item For all $m \in [ M]$, we have $f^* \in \mathcal{F}_M \subset  \mathcal{F}_{M-1} \subset \cdots  \mathcal{F}_1$.
      \item For all $m \in [ M]$, we have $  \frac{2}{3}w_m  \leq   \hat{w}_m  \leq \frac{4}{3}w_m   $.
   \end{enumerate}
   Then with probability $1-\delta$, the event $\Gamma = \bigcap_{i=1}^3 \Gamma_i$ holds.
   
\end{mylem}
The first event is slightly different from that in \cite{CBwithOracle}, 
because in the graph feedback setting, the learner can gain at most $|\mathcal{A}|t$ samples up to round $t$. 
The second and third events come from \cite{CBwithOracle} and \cite{instanceCB_RL}, respectively.
It is straightforward to modify the proof in \cite{CBwithOracle} to derive similar results, 
so we omit their proof here.

Given the high-probability event $\Gamma$, we observe $\hat{f}_m \in \mathcal{F}_m$ and $f^* \in \mathcal{F}_m$ for all $m\in [M]$.
Since $|\mathcal{A}(x;\mathcal{F}_m)|=1$, 
we must have $\pi_f(x) = \pi_{f^*}(x) = \pi_{ \hat{f}_m }(x)$. 
Therefore, for all epoch $m$ and all rounds $t$ in the epoch $m$, we define the following quantities $ \mathcal{R}^{dis}_t(\pi)$ and $ \hat{\mathcal{R}}^{dis}_t(\pi)$. 
$$
\mathcal{R}^{dis}_t(\pi) =  \mathbb{E}_x[ \ind{|\mathcal{A}(x;\mathcal{F}_{m(t)})|>1}    f^*(x,\pi(x))]\text{ and }  \hat{\mathcal{R}}_t^{dis}(\pi) =  \mathbb{E}_x[ \ind{|\mathcal{A}(x;\mathcal{F}_{m(t)})|>1}   \hat{f}_{m(t)}(x,\pi(x))].
$$
We conclude that 
$$
Reg_t(\pi) = \mathcal{R}^{dis}_t (\pi_{f^*} )  -   \mathcal{R}^{dis}_t (\pi )  
$$ 
and 
$$
\widehat{Reg}_t(\pi) = \widehat{\mathcal{R}}^{dis}_t (\pi_{f^*} )  -   \widehat{\mathcal{R}}^{dis}_t (\pi )  
$$ 
when $\Gamma$ holds.

Now we prove a proposition of selected quantities $\lambda_m = \hat{w}_m/ \sqrt{\hat{w}_{m-1}}$.
\begin{myprop}
   \label{prop: decreasing in m}
Assume that $\Gamma$ holds. Then $\rho_m/ \hat{w}_m$ is monotonically non-decreasing in $m$.
\end{myprop}
\begin{myproof}
   We have 
   $
    \frac{\rho_1}{\hat{w}_1} = 0
   $
   for $m=1$ and 
   $$
   \frac{\rho_m}{\hat{w}_m} = \sqrt{  \frac{\eta (\tau_{m-1}-\tau_{m-2})     }{ \hat{w}_m \log( 2\delta^{-1} |\mathcal{F}||\mathcal{A}|  T^2 )   }   }
   $$
   for all $m=2,\cdots,M$.

   Clearly, the property holds for $m=2$.
   Hence, for all $m=3,\cdots,M$, we have 
   $$
   \frac{\rho_{m-1}}{\hat{w}_{m-1}}/ \frac{\rho_m}{\hat{w}_m} = \sqrt{  \frac{ \hat{w}_{m-2}  }{ \hat{w}_{m-1}   }  \frac{ (\tau_{m-2} - \tau_{m-3})  }{  (\tau_{m-1} - \tau_{m-2})  }    } = \sqrt{  \frac{ \hat{w}_{m-2}  }{2 \hat{w}_{m-1}   }   } \leq \sqrt{ \frac{ \frac{4}{3} w_{m-2}   }{ 2\times \frac{2}{3} w_{m-1}  }   } = \sqrt{ \frac{  w_{m-2}   }{   w_{m-1}  }   } \leq 1.
   $$
   The last inequality due to the fact that $\mathcal{F}_M \subset \mathcal{F}_{M-1}\cdots \mathcal{F}_1$ and thus $w_{m-2} = \nu_{m-2} + \mu_{m-2} \leq \nu_{m-1} + \mu_{m-1} \leq  w_{m-1}$.

\end{myproof}

\subsection{Implicit optimization problem}

\begin{myproof}
Let $m$ and $t$ in epoch $m$ be fixed. 
We have 
\begin{align*}
   & \sum_{\pi\in \Psi }  Q_t(\pi)    \widehat{ Reg}_t(\pi) \\
 =  &  \sum_{\pi\in \Psi}  Q_t(\pi) \mathbb{E}_{x_t} \left[  (  \hat{f}_{m}(x_t,\pi_{\hat{f}_{m}}(x_t)) -  \hat{f}_{m}(x_t,\pi(x_t))   )  \right]   \\
 =  &  \mathbb{E}_{x_t} \left[ \sum_{a\in \mathcal{A} }    \sum_{\pi\in \Psi}  \ind{\pi(x_t)=a}   Q_t(\pi) (  \hat{f}_{m}(x_t,\pi_{\hat{f}_{m}}(x_t)) -  \hat{f}_{m}(x_t,a)   ) \right]      \\
 =  & \mathbb{E}_{x_t} \left[   \sum_{a\in \mathcal{A} }   p_t(a|x_t) (  \hat{f}_{m}(x_t,\pi_{\hat{f}_{m}}(x_t)) -  \hat{f}_{m}(x_t,a)   )\right].
\end{align*}
The first and second equalities are the definitions of $\widehat{ Reg}_t(\pi)$ and $Q_t(\pi)$, respectively.

Notice that $\tilde{p}_t$ is a feasible solution to \eqref{eq: sampling probability} and the RHS of the above equations is the object.
Now for the context $x_t$, we have 
\begin{align*}
&   \sum_{a\in \mathcal{A} }  p_t(a|x_t ) (  \hat{f}_{m}(x_t,\pi_{\hat{f}_{m}}(x_t)) -  \hat{f}_{m}(x_t,a)   )       \\
&  \leq \sum_{a\in \mathcal{A} }  \tilde{p}_t(a|x_t ) (  \hat{f}_{m}(x_t,\pi_{\hat{f}_{m}}(x_t)) -  \hat{f}_{m}(x_t,a)   )       \\
& =    \sum_{a\in S_t- \{ \pi_{\hat{f}_{m}}(x_t)  \}   }   \frac{  \hat{f}_{m}(x_t,\pi_{\hat{f}_{m}}(x_t)) -  \hat{f}_{m}(x_t,a) }{ |S_t| + \gamma_t( \hat{f}_{m}(x_t,\pi_{\hat{f}_{m}}(x_t)) -  \hat{f}_{m}(x_t,a)   )        }     \\
 & \leq [ |S_t| -1   ]  / \gamma_t \\
 & \leq \sqrt{ |S_t| - 1 }/  \rho_m.
\end{align*}
The first equality is due to the construction of $p(\cdot|\cdot)$, which is zero of actions outside $S_t$.
Taking expectation over the randomness of $x_t$ and $G_t$, we have 
$$
 \sum_{\pi\in \Psi}  Q_t(\pi) \widehat{ Reg}_t(\pi) \leq   \sqrt{ |S_t| - 1 } /  \rho_m  .
$$
The first inequality is Jensen's inequality and the property of the conditional expectation.

For the second inequality, we first observe that for any policy $\pi \in \Psi$, given any context $x_t \in \mathcal{X}$, 
$$
\frac{1}{q_t(\pi(x_t)|x_t)} \leq  \frac{1}{\tilde{p}_t(\pi(x_t)|x_t)} = |S_t|  + \gamma_t( \hat{f}_{m}(x_t,\pi_{\hat{f}_{m}}(x_t)) -  \hat{f}_{m}(x_t,\pi(x_t))   ) , 
$$
if $\pi(x_t) \in S_t $, and for $\pi(x_t) \notin S_t $, due to the construction of $\tilde{p}_t$ and the linear program constraints, 
there exists an action $a \in S_t$ such that $ \hat{f}_{m}(x_t,\pi_{\hat{f}_{m}}(x_t)) -  \hat{f}_{m}(x_t,\pi(x_t)) \geq \hat{f}_{m}(x_t,\pi_{\hat{f}_{m}}(x_t)) -  \hat{f}_{m}(x_t,a)  $ and $ \pi(x_t) \in \mathcal{N}^{out}_{a}(G_t)$.
Hence, we have
$$
\frac{1}{q_t(\pi(x_t)|x_t)} \leq  \frac{1}{\tilde{p}_t(a |x_t)}  = |S_t|  + \gamma_t( \hat{f}_{m}(x_t,\pi_{\hat{f}_{m}}(x_t)) -  \hat{f}_{m}(x_t,a)   ) \leq  |S_t|  + \gamma_t( \hat{f}_{m}(x_t,\pi_{\hat{f}_{m}}(x_t)) -  \hat{f}_{m}(x_t,\pi(x_t))   ) , 
$$
The second inequality follows immediately by taking expectation over $x_t$ and $G_t$.
\end{myproof}

Compared with IOP in \cite{fasterCB}, 
the key different part is that $ |S_t|$ is replaced by the cardinality $|\mathcal{A}|$ of the whole action set. 
This point highlight the adaptivity to graph feedbacks and show how the graph structure affects the action selection.

Then we need to modify the proof of \cref{lem: IOP} to incorporate the quantity $\nu_m$.

\begin{mylem}
   \label{lem: disagreement-based IOP}
 (Disagreement-based Implicit Optimization Problem). For all epoch $m$ and all rounds $t$ in epoch $m$, $Q_t$ is a feasible solution to the following 
 implicit optimization problem:
 \begin{align}
   &  \mathbb{E} \Big[\sum_{\pi\in \Psi}  Q_t(\pi) \widehat{ Reg}_t(\pi) \Big] \leq  \nu_m \sqrt{\mathbb{E}_G[\alpha(G)] -1}    / \rho_m    \\
   &  \mathbb{E}\left[ \frac{ \ind{ |\mathcal{A}(x;\mathcal{F}_m)| > 1   }  }{q_t(\pi(x)|x)} \right]  \leq \nu_m \mathbb{E}_G[\alpha(G)]  + \sqrt{  \mathbb{E}_G[\alpha(G)] }  \rho_m    \widehat{ Reg}_t(\pi), \forall \pi \in \Psi.
 \end{align}
 
 \end{mylem}

\begin{myproof}
   It can be observed from the construction of $S_t$ that 
   \begin{enumerate}
      \item $\ind{|S_t|>1} \leq  \ind{ |\mathcal{A}(x_t;\mathcal{F}_m)| > 1 } $
      \item $ |S_t| \leq \alpha(G_t)$.
   \end{enumerate}
  Since the exploration set $S_t$ is constructed from the induced subgraph, 
  the cardinality of $S_t$ must be less than or equal to $\mathcal{A}(x_t;\mathcal{F}_m)$.
  The second property holds because $S_t$ is a subset of an independence set of $G_t$.

   From the proof of \cref{lem: IOP}, we know that 
\begin{align*}
   &  \mathbb{E}\Big[\sum_{\pi\in \Psi}  Q_t(\pi) \widehat{ Reg}_t(\pi) \Big]\\
 \leq  & \mathbb{E}[ \sqrt{ |S_t| - 1 } ]/  \rho_m  \\
 =  & (  \mathbb{E}[ \sqrt{ |S_t| - 1 } \ind{ |S_t| > 1 }  ]  + \mathbb{E}[ \sqrt{ |S_t| - 1 } \ind{ |S_t| = 1 }  ] )     /  \rho_m  \\
 =  & \mathbb{E}[ \sqrt{ |S_t| - 1 } \ind{ |S_t| > 1 }  ]    /  \rho_m  \\
 \leq  & \mathbb{E}[ \sqrt{ |S_t| - 1 } \ind{ |\mathcal{A}(x_t;\mathcal{F}_m)| > 1 }  ]    /  \rho_m  \\
 \leq &  \nu_m \sqrt{ \mathbb{E}_{G}[ \alpha(G) ] - 1} /   \rho_m .
\end{align*}
The fourth inequality is due to the fact that $\ind{ |S_t| > 1 } $ implies $\ind{ |\mathcal{A}(x_t;\mathcal{F}_m)| > 1 }$.
The last equality is due to the independence of $x_t$ and $G_t$.

For the second inequality, we have 
\begin{align*}
    & \mathbb{E}\left[ \frac{ \ind{ |\mathcal{A}(x;\mathcal{F}_m)|>1  }  }{q_t(\pi(x)|x)} \right]  \\
  = & \mathbb{E}\left[ \frac{ 1 -  \ind{  |\mathcal{A}(x;\mathcal{F}_m)|=1  }  }{q_t(\pi(x)|x)} \right] \\
  = & \mathbb{E}\left[ \frac{ 1  }{q_t(\pi(x)|x)} \right]   -  \mathbb{E}\left[ \frac{  \ind{  |\mathcal{A}(x;\mathcal{F}_m)|=1   }   }{q_t(\pi(x)|x)} \right]   \\
\leq & \mathbb{E}\left[ \frac{ 1  }{q_t(\pi(x)|x)} \right]  - \prob{   |\mathcal{A}(x;\mathcal{F}_m)|=1    } \\
\leq & \mathbb{E}[ |S_t| ]  +    \sqrt{ |S_t| } \rho_m    \widehat{ Reg}_t(\pi) -  \prob{   |\mathcal{A}(x;\mathcal{F}_m)|=1    } \\
\leq &  \mathbb{E}[ |S_t| \ind{|S_t| = 1   }  ] + \mathbb{E}[ |S_t| \ind{|S_t| > 1   }  ]  +    \sqrt{ \mathbb{E}_{G}[\alpha(G)] } \rho_m    \widehat{ Reg}_t(\pi)-  \prob{   |\mathcal{A}(x;\mathcal{F}_m)|=1    }\\
\leq &  \mathbb{E}[  \ind{|\mathcal{A}(x_t;\mathcal{F}_m)| = 1   }  ] -  \prob{   |\mathcal{A}(x;\mathcal{F}_m)|=1    } + \mathbb{E}_G[ \alpha(G) ] \mathbb{E}[ \ind{|\mathcal{A}(x_t;\mathcal{F}_m)|>1} ] +    \sqrt{ \mathbb{E}_{G}[\alpha(G)] } \rho_m    \widehat{ Reg}_t(\pi)\\
\leq &  \nu_m  \mathbb{E}_{G}[ \alpha(G)  ]  +    \sqrt{ \mathbb{E}_{G}[\alpha(G)] } \rho_m    \widehat{ Reg}_t(\pi) . \\
\end{align*}

\end{myproof}

\subsection{Prediction error}

Our setting do not change the proof procedure of the following lemma \cite{fasterCB}, 
because this lemma does not explicitly involve the number of action set. 
This lemma bounds the prediction error between the true reward and the estimated reward.

\begin{mylem}
   \label{lem: prediction error}
   Assume $\Gamma$ holds. For all epochs $m>1$, all rounds $t$ in epoch $m$, and all policies $\pi \in \Psi$, then 
   $$
   \left|   \widehat{\mathcal{R}}^{dis}_t(\pi) - \mathcal{R}^{dis}_t(\pi) \right| \leq \frac{\lambda_m}{2\rho_{m}} \sqrt{  \max_{1\leq s \leq \tau_{m(t)-1}}   \mathbb{E} \Bigg[  \frac{  \ind{|\mathcal{A}(x;\mathcal{F}_{m(s)})|>1} }{q_{s}(\pi(x)|x)}  \Bigg]     },
   $$
   where the expectation is taken with respect to the randomness of $x$ and $G$.
\end{mylem}

\begin{myproof}
   For any fixed round $t$ and any policy $\pi \in \Psi$, we have 
   $$
   \widehat{\mathcal{R}}_t(\pi) - \mathcal{R}_t(\pi) = \mathbb{E}_{x_t} [  \hat{f}_{m(t)}(x_t,\pi(x_t)) -f^*(x_t,\pi(x_t))  ].
   $$
   For all $s=1,2,\cdots,\tau_{m(t)-1}$, we have 
   \begin{align*}
     & \mathbb{E}_{a_s|x_s}\Bigg[   \sum_{a\in \mathcal{N}_{a_s}(G_s)}  ( \hat{f}_{m(s)}(x_s,a) -f^*(x_s,a) )^2  \Bigg| \mathcal{H}_{s-1}\Bigg] \\
   = & \sum_{ a \in \mathcal{A} } q_{s}(a|x_s)  ( \hat{f}_{m(s)}(x_s,a) -f^*(x_s,a)   )^2 \\ 
   \geq &  q_{s}(\pi(x_s)|x_s)  ( \hat{f}_{m(s)}(x_s,\pi(x_s)) -f^*(x_s,\pi(x_s))   )^2.
   \end{align*}
   Therefore, we have 
   \begin{align*}
      &   \max_{1\leq s \leq \tau_{m(t)-1}}   \mathbb{E} \Bigg[  \frac{  \ind{|\mathcal{A}(x;\mathcal{F}_{m(s)})|>1} }{q_{s}(\pi(x_{s})|x_{s})}  \Bigg]  \sum_{s=\tau_{m(t)-2}+1}^{\tau_{m(t)-1}}  \mathbb{E}_{a_s,x_s} \Bigg[   \sum_{a\in \mathcal{N}_{a_s}(G_s)}  ( \hat{f}_{m(t)}(x_s,a) -f^*(x_s,a) )^2  \Bigg| \mathcal{H}_{s-1} \Bigg] \\
    = & \sum_{s=\tau_{m(t)-2}+1}^{\tau_{m(t)-1}} \mathbb{E} \Bigg[  \frac{ \ind{|\mathcal{A}(x;\mathcal{F}_{m(s)})|>1} }{q_s(\pi(x_s)|x_s)}  \Bigg]   \mathbb{E}_{a_s,x_s} \Bigg[   \sum_{a\in \mathcal{N}_{a_s}(G_s)}  ( \hat{f}_{m(t)}(x_s,a) -f^*(x_s,a) )^2  \Bigg| \mathcal{H}_{s-1} \Bigg] \\
    = & \sum_{s=\tau_{m(t)-2}+1}^{\tau_{m(t)-1}} \mathbb{E}_{x_s}    \Bigg[  \frac{  \ind{|\mathcal{A}(x;\mathcal{F}_{m(s)})|>1} }{q_s(\pi(x_s)|x_s)}  \Bigg]  \mathbb{E}_{x_s} \mathbb{E}_{a_s|x_s} \Bigg[   \sum_{a\in \mathcal{N}_{a_s}(G_s)}  ( \hat{f}_{m(t)}(x_s,a) -f^*(x_s,a) )^2  \Bigg| \mathcal{H}_{s-1} \Bigg] \\
    \geq &  \sum_{s=\tau_{m(t)-2}+1}^{\tau_{m(t)-1}}  \left(  \mathbb{E}_{x_s} \Bigg[   \sqrt{ \mathbb{E}  \Bigg[  \frac{ \ind{|\mathcal{A}(x;\mathcal{F}_{m(s)})|>1} }{q_s(\pi(x_s)|x_s)}  \Bigg]   \mathbb{E}_{a_s|x_s} \Bigg[   \sum_{a\in \mathcal{N}_{a_s}(G_s)}  ( \hat{f}_{m(t)}(x_s,a) -f^*(x_s,a) )^2  \Bigg| \mathcal{H}_{s-1} \Bigg]     }        \Bigg]         \right)^2\\
    \geq &  \sum_{s=\tau_{m(t)-2}+1}^{\tau_{m(t)-1}}  \left(  \mathbb{E}_{x_s} [  \ind{|\mathcal{A}(x;\mathcal{F}_{m(s)})|>1}  | \hat{f}_{m(t)}(x_s,\pi(x_s)) -f^*(x_s,\pi(x_s))   |   ]         \right)^2 \\
    \geq & \sum_{s=\tau_{m(t)-2}+1}^{\tau_{m(t)-1}}  | \widehat{\mathcal{R}}_{m(t)-1}(\pi) - \mathcal{R}_{m(t)-1}(\pi) |^2 \\
    = & (\tau_{m(t)-1} -  \tau_{m(t)-2} )|\widehat{\mathcal{R}}^{dis}_{m(t)-1}(\pi) - \mathcal{R}^{dis}_{m(t)-1}(\pi) |^2.
   \end{align*}
   Here, the inequalities result from Cauchy-Schwarz inequality, the previous deduction, the Jensen's inequality and the convexity of $L_1$ norm, respectively.
   The final equality results from the i.i.d. assumption on context distribution.
Therefore, 
\begin{align*}
   &|\widehat{\mathcal{R}}^{dis}_t(\pi) - \mathcal{R}^{dis}_t(\pi) | \\
   \leq & \sqrt{ \max_{1\leq s \leq \tau_{m-1}}   \mathbb{E} \Bigg[  \frac{ \ind{|\mathcal{A}(x;\mathcal{F}_{m(s)})|>1} }{q_s(\pi(x_s)|x_s)}  \Bigg] \frac{1}{\tau_{m-1} - \tau_{m-2} } \sum_{s=\tau_{m-2}+1}^{\tau_{m-1}}  \mathbb{E}_{a_s,x_s} \Bigg[   \sum_{a\in \mathcal{N}_{a_s}(G_s)}  ( \hat{f}_{m(t)}(x_s,a) -f^*(x_s,a) )^2  \Bigg| \mathcal{H}_{s-1} \Bigg] }.
\end{align*}
We conclude the proof by plugging in the definition of $\Gamma$.
\end{myproof}

The third step is to show that the one-step regret $Reg_t(\pi)$ is close to the one-step estimated regret $\widehat{Reg}_t(\pi)$. 
The following lemma states the result. 
\begin{mylem}
   \label{lem: prediction error of regrets of gap-dependent upper bound}
   Assume $\Gamma$ holds. Let $c_0=4$. 
   For all epochs $m$ and all rounds $t$ in epoch $m$, and all policies $\pi \in \Psi$,
   \begin{align}
      &   Reg_t(\pi)\leq 2  \widehat{Reg}_t(\pi)   + c_0 \hat{w}_m \sqrt{   \mathbb{E}_{G}  [   \alpha(G)  ] } /\rho_m , \label{eq: one-step regret bound 1} \\
      &   \widehat{Reg}_t(\pi)  \leq 2    Reg_t(\pi) + c_0 \hat{w}_m \sqrt{  \mathbb{E}_{G}  [  \alpha(G)  ]  } /\rho_m  \label{eq: one-step regret bound 2}.
   \end{align}
\end{mylem}

\begin{myproof}
We prove this lemma via induction on $m$. 
It is easy to check
$$
Reg_t(\pi) \leq 1, \widehat{Reg}_t(\pi) \leq 1, 
$$
as $\gamma_1 = 0$ and $c_0 \mathbb{E}_{G} \big[   \alpha(G) \big] \geq 1$. Hence, the base case holds.

For the inductive step, fix some epoch $m>1$ and assume that for all epochs $m'<m$, all rounds $t'$ in epoch $m'$, and all $\pi\in\Psi$, 
the inequalities \eqref{eq: one-step regret bound 1} and \eqref{eq: one-step regret bound 2} hold.
We first show that for all rounds $t$ in epoch $m$ and all $\pi\in\Psi$, 
$$
  Reg_t(\pi) \leq 2  \widehat{Reg}_t(\pi)   +c_0 \hat{w}_m \sqrt{   \mathbb{E}_{G}  [   \alpha(G_t)  ] } /\rho_m  .
$$
We have 
\begin{align*}
   & Reg_t(\pi) -   \widehat{Reg}_t(\pi) \\
 =  &   [\mathcal{R}^{dis}_t(\pi_{f^*}  ) -\mathcal{R}^{dis}_t(\pi )]  -    [ \widehat{ \mathcal{R}  }^{dis}_t (\pi_{\hat{f}_m}  ) - \widehat{  \mathcal{R}}^{dis}_t(\pi )   ] \\
 \leq   &  [\mathcal{R}^{dis}_t(\pi_{f^*}  ) -\mathcal{R}^{dis}_t(\pi )]  -   [ \widehat{ \mathcal{R} }^{dis}_t (\pi_{f^*}  ) - \widehat{  \mathcal{R}}^{dis}_t(\pi )   ] \\
 \leq  & | \mathcal{R}^{dis}_t(\pi_{f^*}  )  - \widehat{ \mathcal{R}  }^{dis}_t (\pi_{f^*}  )| +| \mathcal{R}^{dis}_t(\pi ) -  \widehat{  \mathcal{R}}^{dis}_t(\pi )| \\
 \leq  &   \frac{\lambda_m}{2\rho_m} \sqrt{      \max_{1\leq s \leq \tau_{m(t)-1}}   \mathbb{E} \Bigg[  \frac{  \ind{|\mathcal{A}(x;\mathcal{F}_m)|>1} }{q_{s}(\pi_{f^*}(x)|x )} \Bigg]   }  +  \frac{\lambda_m}{2\rho_m} \sqrt{      \max_{1\leq s \leq \tau_{m(t)-1}}   \mathbb{E} \Bigg[  \frac{  \ind{|\mathcal{A}(x;\mathcal{F}_{m(s)})|>1} }{q_{s}(\pi(x)|x )} \Bigg]   } \\
 \leq  &   \frac{  \max_{1\leq s \leq \tau_{m(t)-1}}   \mathbb{E} \Big[  \frac{  \ind{|\mathcal{A}(x;\mathcal{F}_{m(s)})|>1} }{p_{s}(\pi_{f^*}(x_{s})|x_{s},S_{s})} \Big]     }{5\rho_m  \sqrt{   \mathbb{E}_G  [   \alpha(G)  ] }  \hat{w}_{m-1}/\hat{w}_{m}  }          + \frac{   \max_{1\leq s \leq \tau_{m(t)-1}}   \mathbb{E} \Big[  \frac{  \ind{|\mathcal{A}(x;\mathcal{F}_{m(s)})|>1} }{q_{s}(\pi(x )|x  )} \Big]  }{5\rho_m \sqrt{   \mathbb{E}_{G}  [   \alpha(G)  ] }  \hat{w}_{m-1}/\hat{w}_{m}  }      +   \frac{5 \sqrt{   \mathbb{E}_{G}  [   \alpha(G)  ]   } \hat{w}_m   }{8\rho_m  }. 
\end{align*}
The last inequality is by the AM-GM inequality.

From \cref{lem: disagreement-based IOP} we know that 
$$
\max_{1\leq s \leq \tau_{m(t)-1}}   \mathbb{E} \Big[  \frac{  \ind{|\mathcal{A}(x;\mathcal{F}_{m(s)})|>1} }{q_{s}(\pi(x)|x)} \Big]  \leq q_{m-1} \mathbb{E}_{G}[\alpha(G)]   +  \mathbb{E}_{G}[  \sqrt{ \alpha(G) }] \rho_{m-1}  \widehat{ Reg}_t(\pi),
$$
holds for all $ \pi \in \Psi$, for all epoch $m \in [M]$ and for all rounds $t$ in corresponding epochs.
Hence, for epoch $m$ and all rounds $t$ in this epoch, we have
\begin{align*}
   &   \frac{   \max_{1\leq s \leq \tau_{m(t)-1}}   \mathbb{E} \Big[  \frac{  \ind{|\mathcal{A}(x;\mathcal{F}_{m(s)})|>1} }{q_{s}(\pi(x)|x)} \Big]  }{5\rho_m \sqrt{   \mathbb{E}_{G}  [   \alpha(G)  ] }  \hat{w}_{m-1}/\hat{w}_{m}  }   \\
 \leq  &  \frac{ q_{m-1} \mathbb{E}_{G}[\alpha(G)]   +  \mathbb{E}_{G}[  \sqrt{ \alpha(G) }] \rho_{m-1}  \widehat{ Reg}_t(\pi) }{5 \sqrt{  \mathbb{E}_{G}[\alpha(G)]}  \rho_m  \hat{w}_{m-1}/\hat{w}_{m}   },  \text{  (\cref{lem: disagreement-based IOP}) }   \\
 \leq  &  \frac{ q_{m-1} \mathbb{E}_{G}[\alpha(G)]   +  \mathbb{E}_{G}[  \sqrt{ \alpha(G) }] \rho_{m-1}  [ 2 Reg_t(\pi) + c_0  \hat{w}_{m-1}   \sqrt{  \mathbb{E}_{G}  [  \alpha(G)  ]  } /\rho_{m-1}      ]       }{5 \sqrt{  \mathbb{E}_{G}[\alpha(G)]}    \rho_m  \hat{w}_{m-1}/\hat{w}_{m}} , \text{  (inductive assumption) }     \\
 \leq  &  \frac{ q_{m-1} \mathbb{E}_{G}[\alpha(G)]   +  \sqrt{ \mathbb{E}_{G}[  \alpha(G)  ]} \rho_{m-1}  [ 2 Reg_t(\pi) + c_0   \hat{w}_{m-1} \sqrt{  \mathbb{E}_{G}  [  \alpha(G)  ]  } /\rho_{m-1}      ]       }{5 \sqrt{  \mathbb{E}_{G}[\alpha(G)]}    \rho_m  \hat{w}_{m-1}/\hat{w}_{m} },   \text{  (Jensen's inequality) }   \\
 \leq & \frac{2}{5} Reg_t(\pi) \frac{ \rho_{m-1}/\hat{w}_{m-1}  }{ \rho_{m}/\hat{w}_{m}  } + \frac{q_{m-1}+c_0 \hat{w}_{m-1}  }{5\rho_m  \hat{w}_{m-1}/\hat{w}_{m}  }  \sqrt{ \mathbb{E}_{G}[  \alpha(G)  ]}, \text{  (\cref{prop: decreasing in m} and $q_{m-1}\leq w_m$) }   \\
 \leq & \frac{2}{5} Reg_t(\pi)  + \frac{4/3+c_0    }{5    } \hat{w}_{m} \sqrt{ \mathbb{E}_{G}[  \alpha(G)  ]} / \rho_m .  
\end{align*}
We can bound $ \frac{   \max_{1\leq s \leq \tau_{m(t)-1}}   \mathbb{E} \Big[  \frac{  \ind{|\mathcal{A}(x;\mathcal{F}_{m(s)})|>1} }{q_{s}(\pi_{f^*}(x)|x)} \Big]  }{5\rho_m \sqrt{   \mathbb{E}_{G}  [   \alpha(G)  ] }  \hat{w}_{m-1}/\hat{w}_{m}  }    $ in the same way.

Combing all above inequalities yields 
\begin{align*}
Reg_t(\pi) -   \widehat{Reg}_t(\pi) \leq &  \frac{ 2 (4/3+c_0) \hat{w}_m   \sqrt{ \mathbb{E}_{G}[\alpha(G)]}  }{5\rho_m} + \frac{4}{5} \widehat{ Reg}_t(\pi) +    \frac{5 \sqrt{   \mathbb{E}_{G}  [   \alpha(G)  ] } \hat{w}_m    }{ 8\rho_m  } \\
 \leq & \widehat{ Reg}_t(\pi) + (   2 (\frac{4}{3}+ \frac{c_0}{5}) + \frac{5}{8}) \hat{w}_m \frac{ \sqrt{   \mathbb{E}_{G}  [   \alpha(G)  ] }}{ \rho_m  } \\
  \leq & \widehat{ Reg}_t(\pi) + c_0  \frac{ \hat{w}_m \sqrt{   \mathbb{E}_{G}  [   \alpha(G)  ] }}{ \rho_m  }.
\end{align*}

Similarly, we have
\begin{align*}
   &  \widehat{Reg}_t(\pi) -Reg_t(\pi) \\
   =  &     [ \widehat{ \mathcal{R}  }_t^{dis} (\pi_{\hat{f}_m}  ) - \widehat{  \mathcal{R}}_t^{dis}(\pi )   ] -  [\mathcal{R}^{dis}_t(\pi_{f^*}  ) -\mathcal{R}^{dis}_t(\pi )] \\
   \leq   &     [ \widehat{ \mathcal{R}  }_t^{dis} (\pi_{\hat{f}_m}  ) - \widehat{  \mathcal{R}}^{dis}_t(\pi )   ] - [\mathcal{R}^{dis}_t(\pi_{\hat{f}_m}  ) -\mathcal{R}^{dis}_t(\pi )]\\
   \leq  & | \mathcal{R}^{dis}_t(\pi_{\hat{f}_m}  ) - \widehat{ \mathcal{R}  }_t^{dis} (\pi_{\hat{f}_m}  )| +| \mathcal{R}^{dis}_t(\pi ) -  \widehat{  \mathcal{R}}_t^{dis}(\pi )|.
\end{align*}
We can bound the above terms in the same steps.

\end{myproof}

\subsection{Incorporating the uniform gap}

\begin{mylem}
   Assume $\Gamma$ holds. For all epochs $m$ and all rounds $t$ in epoch $m$, and all $f \in \mathcal{F}_m$, we have 
   $$
   Reg_t(\pi_f) \leq  6 \hat{w}_m  \sqrt{   \mathbb{E}_{G}  [   \alpha(G)  ] } M /\rho_m.
   $$
\end{mylem}

\begin{myproof}
   We rewrite $Reg_t(\pi_f)$ as $\mathbb{E}[ \ind{\pi_f(x) \neq \pi_{f^*}} ( f^*(x, \pi_{f^*}(x)  )  -  f^*(x, \pi_{f}(x)  )  )  ]$.
   Hence we have 
   \begin{align*}
      & Reg_t(\pi_f) \\
      = & \mathbb{E}[ \ind{\pi_f(x) \neq \pi_{f^*}(x)} ( f^*(x, \pi_{f^*}(x)  )  -  f^*(x, \pi_{f}(x)  )  )  ] \\
      = & \mathbb{E}\Big[ \ind{\pi_f(x) \neq \pi_{f^*}(x)} ( f^*(x, \pi_{f^*}(x)  )  -   f(x, \pi_{f^*}(x)  ) + f(x, \pi_{f^*}(x)  )  -  f(x, \pi_{f}(x)  ) \\ 
        & \ \ \  + f(x, \pi_{f}(x)  ) - f^*(x, \pi_{f}(x)  )  )  \Big] \\
      \leq  &  \mathbb{E}[ \ind{\pi_f(x) \neq \pi_{f^*}(x)}  (|f^*(x, \pi_{f^*}(x)  )  -  f(x, \pi_{f^*}(x))|  +  | f^*(x, \pi_{f}(x) )  -  f(x, \pi_{f}(x)) |  )      ].
   \end{align*}
   We now consider the term $(Reg_t(\pi_f))^2$ as following. 
   \begin{align*}
      & (Reg_t(\pi_f))^2 \\
      \leq & (\mathbb{E}[ \ind{\pi_f(x) \neq \pi_{f^*}(x)}  (|f^*(x, \pi_{f^*}(x)  )  -  f(x, \pi_{f^*}(x))|  +  | f^*(x, \pi_{f}(x) )  -  f(x, \pi_{f}(x)) |  )      ] )^2\\
      \leq  &  \mathbb{E}\left[  \bigg(\frac{\ind{ |\mathcal{A}(x;\mathcal{F}_m)| > 1  } }{  q_{t-1}(\pi_f(x)|x) } +  \frac{\ind{ |\mathcal{A}(x;\mathcal{F}_m)| > 1  } }{  q_{t-1}(\pi_{f^*}(x)|x) } \bigg)  \mathbb{E}_{x,a\sim p_{t-1}(\cdot|x )}  (f^*(x, a  )  -  f(x,a) )^2      \right] \\
      \leq  &  \mathbb{E}\left[  \bigg(\frac{\ind{ |\mathcal{A}(x;\mathcal{F}_m)| > 1  } }{  q_{t-1}(\pi_f(x)|x) } +  \frac{\ind{ |\mathcal{A}(x;\mathcal{F}_m)| > 1  } }{  q_{t-1}(\pi_{f^*}(x)|x) } \bigg)   \right]  \frac{  (2\beta_m +C_{\delta})    }{ n_{m}/2 }   \\
      \leq  &  \mathbb{E}\left[  \bigg(\frac{\ind{ |\mathcal{A}(x;\mathcal{F}_m)| > 1  } }{  q_{t-1}(\pi_f(x)|x) } +  \frac{\ind{ |\mathcal{A}(x;\mathcal{F}_m)| > 1  } }{  q_{t-1}(\pi_{f^*}(x)|x) } \bigg)   \right]  \frac{  2M\eta \lambda_m^2   }{ \rho_m^2   } .
   \end{align*}

From \cref{lem: disagreement-based IOP} we know 
\begin{align*}
 &  \mathbb{E}\left[  \frac{\ind{ |\mathcal{A}(x;\mathcal{F}_m)| > 1  } }{  q_{t-1}(\pi_f(x)|x) }    \right]      \\
\leq & q_{m-1} \mathbb{E}_{G}[\alpha(G)] +  \sqrt{ \mathbb{E}[\alpha(G)]  } \rho_{m-1} \widehat{Reg}_{\tau_{m-1}}(\pi_f) \\
\leq & q_{m-1} \mathbb{E}_{G}[\alpha(G)] + 2\sqrt{ \mathbb{E}[\alpha(G)]  } \rho_{m-1}  {Reg}_{\tau_{m-1}}(\pi_f) + c_0\hat{w}_{m-1} \mathbb{E}[\alpha(G)]  \\
\leq & \frac{3}{2}\hat{w}_{m-1} \mathbb{E}_{G}[\alpha(G)] + 2\sqrt{ \mathbb{E}[\alpha(G)]  } \rho_{m-1}  {Reg}_{\tau_{m-1}}(\pi_f) + c_0\hat{w}_{m-1} \mathbb{E}[\alpha(G)].
\end{align*}  
and 
\begin{align*}
   &  \mathbb{E}\left[  \frac{\ind{ |\mathcal{A}(x;\mathcal{F}_m)| > 1  } }{  q_{t-1}(\pi_{f^*}(x)|x) }    \right]      \\
  \leq & q_{m-1} \mathbb{E}_{G}[\alpha(G)] + \sqrt{ \mathbb{E}[\alpha(G)]  } \rho_{m-1} \widehat{Reg}_{\tau_{m-1}}(\pi_{f^*}) \\
  \leq & q_{m-1} \mathbb{E}_{G}[\alpha(G)] + 2\sqrt{ \mathbb{E}[\alpha(G)]  } \rho_{m-1} {Reg}_{\tau_{m-1}}(\pi_{f^*}) + c_0 \hat{w}_{m-1} \mathbb{E}[\alpha(G)]  \\
  \leq & \frac{3}{2}\hat{w}_{m-1} \mathbb{E}_{G}[\alpha(G)] + 72 \hat{w}_{m-1} \mathbb{E}_{G}[\alpha(G)].
  \end{align*} 
Plugging the above two inequalities yields 
$$
(Reg_t(\pi_f))^2 \leq (  2 \sqrt{\mathbb{E}_{G}[\alpha(G)]} \rho_{m-1}  Reg_t(\pi_f)  + (2c_0+3) \hat{w}_{m-1} \mathbb{E}_{G}[\alpha(G)]       ) \frac{  2M\eta \hat{w}_m^2    }{ \hat{w}_{m-1} \rho_m^2   } .
$$
which implies 
$$
(Reg_t(\pi_f))^2 \leq (  2 \sqrt{\mathbb{E}_{G}[\alpha(G)]} \frac{\rho_m \hat{w}_{m-1}}{ \hat{w}_m}  Reg_t(\pi_f)  + (2c_0+3) \hat{w}_{m-1} \mathbb{E}_{G}[\alpha(G)]       ) \frac{  2M\eta \hat{w}_m^2    }{ \hat{w}_{m-1} \rho_m^2   } 
$$
according to \cref{prop: decreasing in m}.

Solving the inequality for $Reg_t(\pi_f)$ shows that 
$$
Reg_t(\pi_f) \leq c_1  \hat{w}_m  \log T  \sqrt{ \mathbb{E}_{G}[\alpha(G)]} /\rho_m,
$$
where $c_1 = 6$.

\end{myproof}

At this point, we can bound the regret within each epoch using the above, 
which gives a bound in terms of the empirical disagreement probability $\hat{w}_m$. 
To proceed, we relate this quantity to the policy disagreement coefficient.
Our proof can directly follow from that in \cite{instanceCB_RL} by replacing 
$ A$ with $\mathbb{E}_{G}[\alpha(G)]$ and $A/\gamma_m$ with $ \sqrt{ \mathbb{E}_{G}[\alpha(G)]} /\rho_m $.
Our regret analysis builds on the framework in \cite{fasterCB}.
Hence, we can directly write down the following lemma. 
\begin{mylem}
   \label{lem: gap-dependent upper bound}
   Assume that $\Gamma$ holds. For any fix $\epsilon >0$ and every $m\in [M]$, we have 
   $$
   \sum_{\pi\in\Psi} Q_t(\pi) Reg_t(\pi) \leq \max \left\{ \epsilon,  \theta^{csc}(\mathcal{F},\epsilon)  \frac{c_2  \mathbb{E}[ \alpha(G)  ]  \log(2\delta^{-1}T^2|\mathcal{F}|)     }{ n_{m-1}  }  \right\} + \frac{256\log(4M/\delta)}{ n_{m-1}},
   $$
   where $\theta^{csc}(\mathcal{F},\epsilon) =  \sup_{\epsilon \geq \epsilon_0} \frac{1}{\epsilon} \mathbb{P}_{\mathcal{X}}(x \in \mathcal{X}: \exists f \in \mathcal{F}_{\epsilon}^{csc} \text{ such that } \pi_f(x)\neq \pi_{f^*}(x)   ) $ and 
   $ \mathcal{F}_{\epsilon}^{csc} = \{  f\in\mathcal{F}|  \mathcal{R}(\pi_{f^*}) -  \mathcal{R}(\pi_{f}) \leq \epsilon   \}  $.
\end{mylem}

\begin{mycol}
   Assume that $\Gamma$ and \cref{asp: uniform gap} holds. For any fix $\epsilon >0$ and every $m\in [M]$, we have 
   $$
   \sum_{\pi\in\Psi} Q_t(\pi) Reg_t(\pi) \leq \max \left\{ \epsilon \Delta,   \frac{c_2 \theta^{pol}(\mathcal{F},\epsilon)  \mathbb{E}[ \alpha(G)  ]  \log(2\delta^{-1}T^2|\mathcal{F}|)     }{ \Delta n_{m-1}  }  \right\} + \frac{256\log(4M/\delta)}{ n_{m-1}}.
   $$
\end{mycol}
  
\begin{myproof}
   We replace $\epsilon \Delta$ with $\epsilon$ in \cref{lem: gap-dependent upper bound}.
   Since
   $$
   \mathcal{F}_{\epsilon \Delta}^{csc} = \{  f\in\mathcal{F}|  \mathcal{R}(\pi_{f^*}) -  \mathcal{R}(\pi_{f}) \leq \epsilon \Delta  \},
   $$
   we have
   $$
   \mathbb{P}_{\mathcal{X}}(x \in \mathcal{X}: \exists f \in \mathcal{F}_{\epsilon} \text{ such that } \pi_f(x)\neq \pi_{f^*}(x)   ) \Delta \leq  \mathcal{R}(\pi_{f^*}) -  \mathcal{R}(\pi_{f}) \leq \epsilon \Delta.
   $$
   We conclude that 
   $$
   \mathbb{P}_{\mathcal{X}}(x \in \mathcal{X}: \exists f \in \mathcal{F}_{\epsilon} \text{ such that } \pi_f(x)\neq \pi_{f^*}(x)   ) \leq \epsilon 
   $$
   and thus 
   $$
   \theta^{csc}(\mathcal{F},\epsilon \Delta) \leq \sup_{\epsilon \geq \epsilon' } \frac{\mathbb{P}_{\mathcal{X}}(x \in \mathcal{X}: \exists f \in \mathcal{F}_{\epsilon'} \text{ such that } \pi_f(x)\neq \pi_{f^*}(x)   )}{ \epsilon' \Delta  } = \theta^{pol}(\mathcal{F},\epsilon)/ \Delta.
   $$

\end{myproof}

Now it is time to prove the theorem of the gap-dependent upper bound.

\begin{myproof}
   Our regret analysis builds on the framework in \cite{fasterCB}.

   \textbf{Step 1:} proving an implicit optimization problem for $Q_t$ in \cref{lem: IOP}.

   \textbf{Step 2:} bounding the prediction error between $ \widehat{\mathcal{R}}^{dis}_t(\pi)$ and $\mathcal{R}^{dis}_t(\pi)$ in \cref{lem: prediction error}. 
   Then we can show that the one-step regrets $\widehat{Reg}_t(\pi)$ and $ {Reg}(\pi) $ are close to each other.
   
   \textbf{Step 3:} bounding the cumulative regret $Reg(T)$. 

   By Lemma 4 of \cite{fasterCB}, we know
   $$
   \mean{Reg(T)} = \sum_{t=1}^T \sum_{ \pi \in \Psi} Q_{t}(\pi) Reg(\pi) .
   $$
   Combing \cref{lem: gap-dependent upper bound}, we know 
   \begin{align*}
     \mathbb{E}[Reg(T)] =&   \sum_{t=1}^T \sum_{ \pi \in \Psi} Q_{m(t)}(\pi) Reg_t(\pi) \\
      \leq & \sum_{m=1}^M \sum_{t=\tau_{m-1}+1}^{\tau_m}  \sum_{ \pi \in \Psi} Q_{t}(\pi) Reg_t(\pi) \\
      \leq & \sum_{m=1}^M n_{m}  \left(   \max\{  \epsilon\Delta ,  \frac{c_2 \theta^{pol}(\mathcal{F},\epsilon)  \mathbb{E}[ \alpha(G)  ]  \log(2\delta^{-1}T^2|\mathcal{F}|)     }{ \Delta  n_{m-1} }    \} +   \frac{ 256\log(4M/\delta) }{  n_{m-1}}      \right)      \\
      \leq & \sum_{m=1}^M   \left(   \max\{  \epsilon\Delta n_{m},  \frac{2c_2 \theta^{pol}(\mathcal{F},\epsilon)  \mathbb{E}[ \alpha(G)  ]  \log(2\delta^{-1}T^2|\mathcal{F}|)     }{ \Delta  }    \}   +  512 \log(4M/\delta)     \right)\\
      \leq &  \max\left\{ \epsilon\Delta T,    \frac{2c_2 \theta^{pol}(\mathcal{F},\epsilon)  \mathbb{E}[ \alpha(G)  ]  \log(2\delta^{-1}T^2|\mathcal{F}|) \log T    }{ \Delta  }        \right\} + 512 \log T \log(4\log T/\delta) 
   \end{align*}
\end{myproof}

\subsection{Minimax optimality}

It can be observed from the definition that $\theta^{pol}(\mathcal{F},\epsilon) \leq \frac{1}{\epsilon}$.
We choose $\epsilon = \tilde{\Theta}( \frac{1}{\Delta \sqrt{T}}  )$ so that the minimax result scale with 
$\tilde{\mathcal{O}}(\sqrt{\mathbb{E}[ \alpha(G)  ] T \log |\mathcal{F}|   }  )$.

 \section{Additional Algorithms}
 \label{sec in appendix: algorithms}

 \begin{algorithm}[h]
	\renewcommand{\algorithmicrequire}{\textbf{Input:}}
	\renewcommand{\algorithmicensure}{\textbf{Output:}}
	\caption{ConstructExplorationSet}
	\label{alg: ConstructExplorationSet}
	\begin{algorithmic}[1]
      \Require the adjacency matrix $G_t$, the adaptive set $\mathcal{A}(x_t;\mathcal{F}_m)$, gaps $\Delta_{a,t}$ for $a\in\mathcal{A}(x_t;\mathcal{F}_m)$
      \State Sort gaps in the ascending order 
      \State Initialize auxiliary set $B_t$ and the exploration set $S_t$ to be empty
      \For { each arm $a$ in $\mathcal{A}(x_t;\mathcal{F}_m)$}
         \If {  the arm $a$ is not in $B_t$  }
         \State Put the arm $a$ in $\mathcal{A}(x_t;\mathcal{F}_m)$
         \State Update $B_t= B_t \cup \mathcal{N}^{out}_a(G_t)$ 
         \EndIf
      \EndFor
      \Ensure An exploration set $S_t$
   \end{algorithmic}  
\end{algorithm}

\begin{algorithm}[h]
	\renewcommand{\algorithmicrequire}{\textbf{Input:}}
	\renewcommand{\algorithmicensure}{\textbf{Output:}}
	\caption{A random graph generator}
	\label{alg: random graph generator}
	\begin{algorithmic}[1]
      \Require The number of nodes $K$, a dense factor $\eta$
      \State Initialization: a $K\times K$ identity matrix $G$, a counter $t$
      \Repeat 
         \State Uniformly sample two nodes $u,v$ in $[K]$
         \State Add the edge $(u,v)$ and $(v,u)$, i.e., $G[u][v] = G[v][u] = 1$
         \State $t = t+1$
      \Until{ $t\geq \eta \times K^2$}
      \Ensure An adjacency matrix $G$
   \end{algorithmic}  
\end{algorithm}